%% file: main.tex
\documentclass[10pt,twocolumn,letterpaper]{article}

\usepackage[pagenumbers]{cvpr} 

\input{preamble}
\definecolor{cvprblue}{rgb}{0.21,0.49,0.74}
\definecolor{skillblue}{RGB}{100,160,200}
\usepackage[pagebackref,breaklinks,colorlinks,allcolors=cvprblue]{hyperref}
\usepackage{xcolor}

\def\confName{CVPR}
\def\confYear{2027}

\usepackage{fontawesome5} 

\title{
\textcolor{skillblue}{SkillEarth}: Teaching General-Purpose Agents to Act on \textcolor{skillblue}{Earth} via \textcolor{skillblue}{Skill} Evolution
}

\author{
Haojun Tang$^{1,2,\dagger}$ \hspace{0.3em}
Wenda Zhao$^{2,}$ \textsuperscript{\faEnvelope} \hspace{0.3em}
Xiang Zhou$^{2}$ \hspace{0.3em}
Qiwei Lin$^{3}$ \hspace{0.3em}
Lida Huang$^{4,5}$ \hspace{0.3em}
Yan Wang$^{1,}$ \textsuperscript{\faEnvelope}
\\[0.5em]
$^{1}$Institute for AI Industry Research (AIR), Tsinghua University \\
$^{2}$Dalian University of Technology \\
$^{3}$Beijing Institute of Radio Measurement \\
$^{4}$School of Safety Science, Tsinghua University \\
$^{5}$Hefei Institute for Public Safety Research, Tsinghua University \\
{\tt\normalsize
tanghaojun@mail.dlut.edu.cn \hspace{0.3em}
zhaowenda@dlut.edu.cn \hspace{0.3em}
wangyan@air.tsinghua.edu.cn
}
\\[0.3em]
{\large
\href{https://xinr-tang.github.io/SkillEarth-homepage/}{
    \faGithub\hspace{0.35em}
    \textbf{https://xinr-tang.github.io/SkillEarth-homepage/}
}
}
}

\usepackage{makecell}
\usepackage{booktabs} 
\usepackage{pifont}   
\usepackage[table]{xcolor}
\usepackage{multirow}
\usepackage{xspace}
\newcommand{\name}{SkillEarth\xspace}

\newcommand{\rankone}[1]{\colorbox{red!12}{#1}}
\newcommand{\ranktwo}[1]{\colorbox{cyan!12}{#1}}
\newcommand{\rankthree}[1]{\colorbox{green!12}{#1}}
\newcommand{\venue}[2]{{\scriptsize (#1)~\cite{#2}}}
\usepackage{cuted}
\usepackage{caption}

\usepackage{algorithm}
\usepackage{algpseudocode}
\algrenewcommand\algorithmicrequire{\textbf{Input:}}
\algrenewcommand\algorithmicensure{\textbf{Output:}}
\algtext*{EndIf}

\usepackage[most]{tcolorbox}
\usepackage{listings}

\tcbset{
    promptstyle/.style={
        enhanced,
        breakable,
        listing only,
        colback=gray!3,
        colframe=black!35,
        colbacktitle=gray!15,
        coltitle=black,
        fonttitle=\bfseries\small,
        boxrule=0.4pt,
        arc=1pt,
        left=5pt,
        right=5pt,
        top=5pt,
        bottom=5pt,
        listing options={
            basicstyle=\ttfamily\scriptsize,
            breaklines=true,
            breakatwhitespace=true,
            breakindent=0pt,
            breakautoindent=false,
            columns=fullflexible,
            keepspaces=true,
            showstringspaces=false,
            tabsize=2
        }
    }
}

\begin{document}
\twocolumn[{%
\renewcommand\twocolumn[1][]{#1}%
\maketitle
\begin{center}
    \centering
    \includegraphics[width=\textwidth]{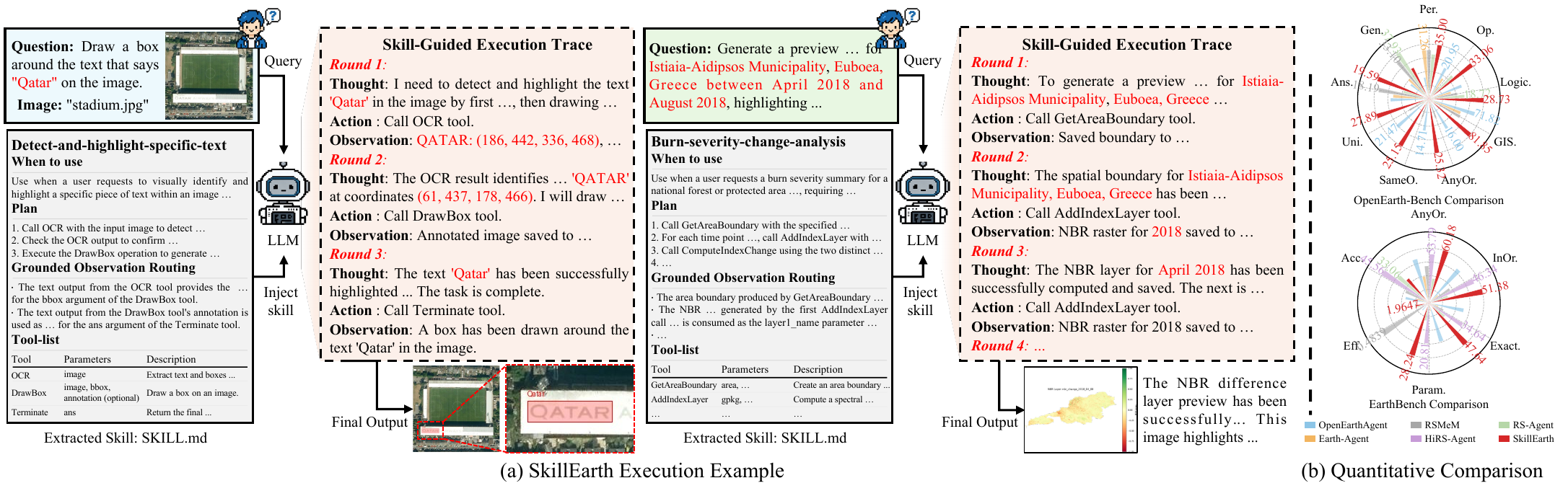}
    \captionof{figure}{Teaser of SkillEarth. (a) Skill-guided execution examples showing how extracted skills guide multi-step tool execution. (b) Quantitative comparisons with existing RS agents on OpenEarth-Bench~\cite{OpenEarthAgentSFT} (End-to-end mode) and Earth--Bench~\cite{EarthAgentEO} (IF mode).}
    \label{Teaser}
\end{center}%
}]

\begingroup
\makeatletter
\renewcommand{\@makefntext}[1]{\noindent #1}
\makeatother
\renewcommand{\thefootnote}{}
\footnotetext{%
    \begin{tabular}[t]{@{}l@{}}
        \faEnvelope\hspace{0.2em} Corresponding authors. \\[2pt]
        $\dagger$ Work done by Haojun Tang as a student researcher at AIR.
    \end{tabular}
}
\endgroup

\input{sec/0_abstract}
\input{sec/1_intro}
\input{sec/2_related_work}

\input{sec/3_method}

\input{sec/4_experiments}
\input{sec/5_conclusion}
\input{sec/6_statement}
{
    \small
    \bibliographystyle{ieeenat_fullname}
    \bibliography{main}
}

\input{sec/X_suppl}

\end{document}

%% file: sec/0_abstract.tex
\begin{abstract}
Recently developed remote sensing (RS) agents execute real-world tasks with complex workflows through multi-step tool invocation.
These agents, however, typically rely on predefined expertise or parameter updates. 
They thus require updating expert knowledge or retraining models whenever new tools, tasks, or execution patterns emerge, limiting their continual adaptability in dynamic Earth environments.
In this work, we seek to equip general-purpose agents with RS tool-call knowledge that can be automatically extracted from demonstrated trajectories without expert intervention or parameter updates.
Specifically, we propose \name, a self-evolving skill extraction method that externalizes RS tool-call knowledge as reusable, retrievable, and evolvable skill files and refines them through observation-grounded evolution.
Unlike existing automatic skill extraction methods that rely on outcome-based reflection to refine skills holistically, \name grounds skill evolution in real tool-chain execution and combines dual-judge feedback with a grounded observation routing graph for precise component-level refinement.
The resulting skills are generalizable, reliable, and transferable: they improve overall RS tool-call performance across three frontier LLMs and four open-source LLMs, are refined through real tool execution feedback, and support cross-model reuse without retraining.
Experiments on OpenEarth-Bench and Earth-Bench demonstrate that \name achieves better tool selection, argument correctness, execution-order consistency, and task completion than existing RS agents and skill extraction methods.
\end{abstract}

%% file: sec/1_intro.tex
\section{Introduction}
Real-world remote sensing (RS) tasks typically rely on complex workflows that coordinate multiple interdependent steps. 
For example, disaster response often involves mapping damage, assessing affected areas, and then delivering actionable information~\cite{disaster1}.
In practice, these steps require collaboration across multiple departments, making the overall workflow time-consuming and inefficient~\cite{integration}. 
Therefore, how to simplify such cumbersome workflows and enable automated task execution remains an important question.

Against this backdrop, recent studies explore multi-modal large language models (MLLMs) for automating RS tasks.
Early efforts focus on fine-tuning RS-specialized MLLMs for image interpretation~\cite{RSGPT, EarthGPT}, visual question answering~\cite{SkyEyeGPT, EarthGPT}, and scene understanding~\cite{GeoChat, LHRSBot}.
Despite their effectiveness, these methods support only a few predefined tasks with single-step execution.
More recently, tool-augmented RS agents have emerged to automate complex, multi-step workflows by leveraging diverse RS tools.
A central challenge in developing such agents is how to equip MLLMs with the domain knowledge required for reliable RS tool use.
To address this, recent methods explore expert knowledge~\cite{RSAgent,CangLingKnowFlow,EarthAgentHTAM}, external memory~\cite{GeoEvolver,RSMeM}, or model fine-tuning~\cite{OpenEarthAgentSFT,HiRS-Agent}.
However, adapting these methods to new tools and tasks requires continual expert updates, memory maintenance, or model retraining, making them difficult to adapt efficiently to dynamic earth observation environments.
This raises the central question of this work: \textbf{\textit{Can a general-purpose agent acquire reliable, reusable, and transferable RS tool-call knowledge without model retraining or expert intervention?}}

\begin{figure}[t]
    \centering
    \includegraphics[width=0.93\linewidth]{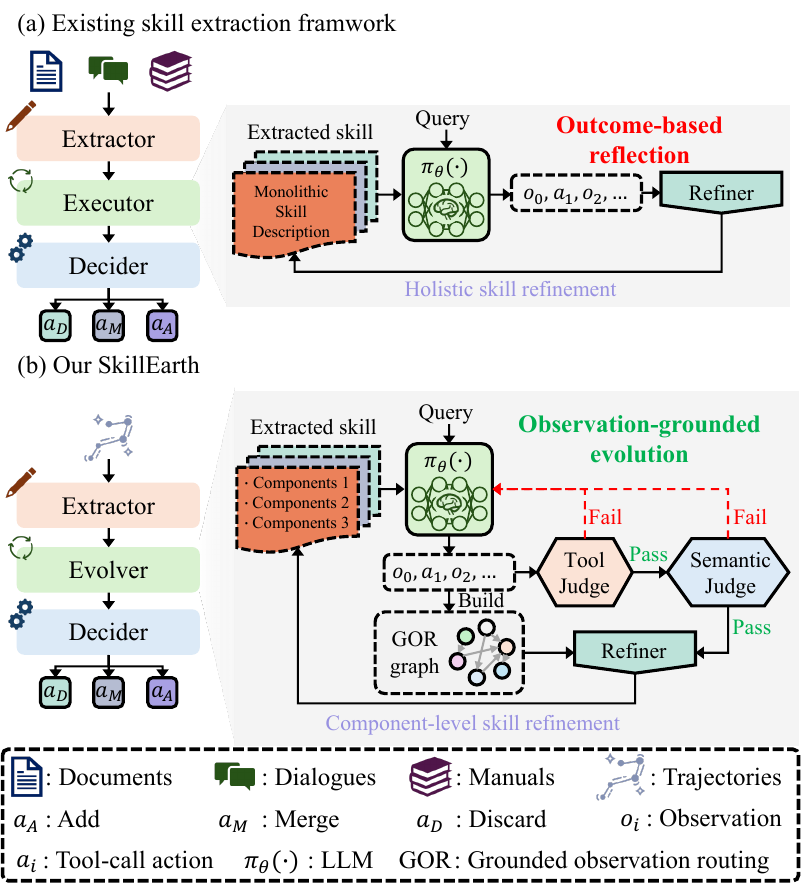}
    \caption{Comparison of (a) existing skill extraction with outcome-based holistic refinement and (b) \name with component-level refinement via observation-grounded evolution.}
    \label{fig:motivation}
\end{figure}

In this paper, we externalize RS tool-call knowledge as textual files, termed skills, that general-purpose agents can retrieve and execute.
Recent studies extract such skills from documents, dialogues, or manuals to reduce expert authoring~\cite{Anything2Skill,AutoSkill,SkillX,Trace2Skill,SkillRevise}.
As shown in Fig.~\ref{fig:motivation}(a), some methods~\cite{SkillX,Trace2Skill,SkillRevise} follow an extract-execute-decide paradigm: an \emph{Extractor} distills candidate skills, an \emph{Executor} refines them through execution and reflection, and a \emph{Decider} adds, merges, or discards them.
However, these methods rely on outcome-based reflection, which provides limited evidence about where failures occur and why.
Moreover, holistic refinement makes skill revision coarse-grained and hard to control.
This limits their use in RS workflows, where a skill must specify tool selection, required and optional arguments, and, crucially, how outputs from one tool bind to inputs of another.
We therefore advance outcome-based reflection into observation-grounded evolution.
As shown in Fig.~\ref{fig:motivation}(b), we represent each skill as multiple independently evolvable components and replace the conventional \emph{Executor} with an \emph{Evolver}.
The \emph{Evolver} applies the extracted skill to its corresponding query, while a \emph{tool judge} verifies tool calls and a \emph{semantic judge} evaluates the final answer.
For each successful trajectory, it further builds a grounded observation routing (GOR) graph, whose nodes denote tool invocations and edges capture how tool outputs bind to arguments of downstream tools.
Using the execution results and GOR graph, the \emph{Refiner} localizes evidence to relevant components and selectively revises them, turning holistic rewriting into controllable component-level evolution.

\begin{figure*}[t]
    \centering
    \includegraphics[width=\textwidth]{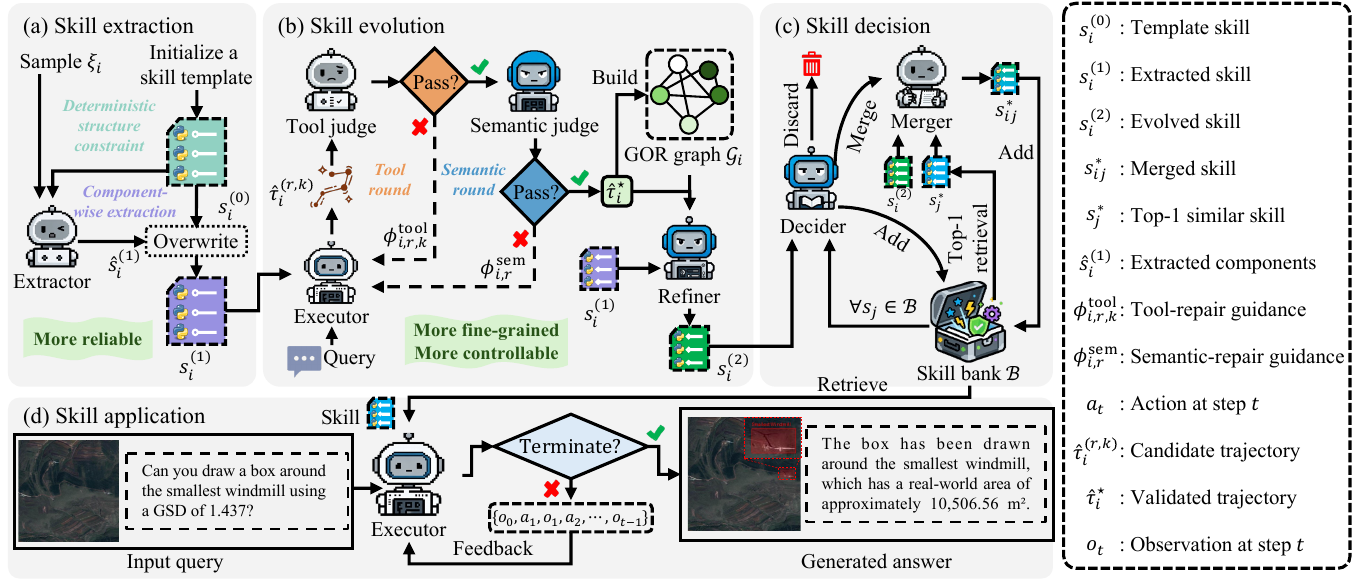}
    \caption{Overview of SkillEarth. (a) The \emph{Extractor} constructs component-wise skills by overwriting templates with LLM-inferred components. (b) The \emph{Evolver} validates and refines skills using dual-judge feedback and GOR graphs. (c) The \emph{Decider} merges, discards, or adds skills to the skill bank. (d) During application, the \emph{Executor} uses retrieved skills to generate final answers.}
    \label{fig:overview}
\end{figure*}

To this end, we propose \name, a self-evolving skill extraction method that automatically extracts RS tool-call skills from demonstrated trajectories and refines them based on execution-grounded feedback.
Specifically, the \emph{Extractor} independently extracts each skill component.
The \emph{Evolver} iteratively executes the extracted skill in a real tool environment, incorporating error diagnoses from tool and semantic judges (e.g., incorrect capitalization or unsupported tool aliases) into the LLM's system prompt to guide next attempts.
Once a valid trajectory is obtained, \name constructs a GOR graph to capture observation routing across tool calls.
The \emph{Refiner} then leverages this graph and execution feedback to identify and revise corresponding components.
Finally, the \emph{Decider} determines whether to add, merge, or discard each skill based on its semantics, task type, and tool-call details.
The resulting skills generalize reliably across different LLMs (three frontier and four open-source models), improving RS task performance without parameter updates.
They are also transferable across models: skills extracted by one LLM can guide another LLM.
Fig.~\ref{Teaser} illustrates how the extracted skills guide multi-step tool execution and highlights the superior End-to-End performance of \name compared with existing methods.

Our contributions are summarized as follows:
\begin{itemize}
    \item We propose \name, a self-evolving skill extraction method that automatically extracts and evolves reusable RS tool-call skills from demonstrated trajectories without expert intervention or parameter updates.
    
    \item We propose the observation-grounded evolution, integrating dual-judge feedback and GOR graphs for precise failure localization and targeted component-level refinement.
    
    \item Extensive experiments on OpenEarth-Bench~\cite{OpenEarthAgentSFT} and Earth-Bench~\cite{EarthAgentEO} demonstrate the generalizability, reliability, and cross-model transferability of the extracted skills.
\end{itemize}

%% file: sec/2_related_work.tex
\section{Related Work}
\label{sec:related_work}

\subsection{MLLM-Based Remote Sensing Models}
Early RS models relied on task-specific architectures for image classification~\cite{classification}, object detection~\cite{detection} and image generation~\cite{mmdiff, CPD}. 
Recent studies employ MLLMs as unified frameworks for RS vision-language understanding, including visual question answering~\cite{RSGPT}, multi-sensor understanding~\cite{EarthGPT}, scene understanding~\cite{GeoChat}, and instruction following~\cite{SkyEyeGPT}. 
However, these models mainly support predefined tasks with single-step inference and cannot coordinate heterogeneous operations required by complex RS workflows. 
In contrast, \name encodes RS tool-calling knowledge into skill files, enabling general-purpose agents to execute multi-step workflows involving perception, spectral analysis, and GIS processing.

\subsection{Tool-Augmented Remote Sensing Agents}
Tool-augmented agents use external tools for multi-step reasoning and execution~\cite{ReAct,RestGPT}.
Building on this paradigm, existing RS agents incorporate expert knowledge for tool planning~\cite{RSAgent,CangLingKnowFlow}, memory for experience refinement~\cite{GeoEvolver,RSMeM}, or supervised fine-tuning on execution trajectories~\cite{OpenEarthAgentSFT}.
Several benchmarks evaluate their capabilities, including ThinkGeo~\cite{ThinkGeo} for spatial reasoning and tool use, Earth-Bench~\cite{EarthAgentEO} for cross-modal tasks under planning and instruction-following settings, GeoMMBench~\cite{GeoMMAgent} for multimodal geoscience reasoning, and OpenEarth-Bench~\cite{OpenEarthAgentSFT} for large-scale tool-call evaluation.
Despite these advances, adapting existing RS agents to new tools, tasks, or execution patterns requires expert maintenance, memory updates, or model retraining.
In contrast, \name externalizes RS tool-call knowledge as evolvable skill files that can be automatically extracted, selectively refined, and transferred across LLMs without fine-tuning.

\begin{figure*}[t]
    \centering
    \includegraphics[width=\textwidth]{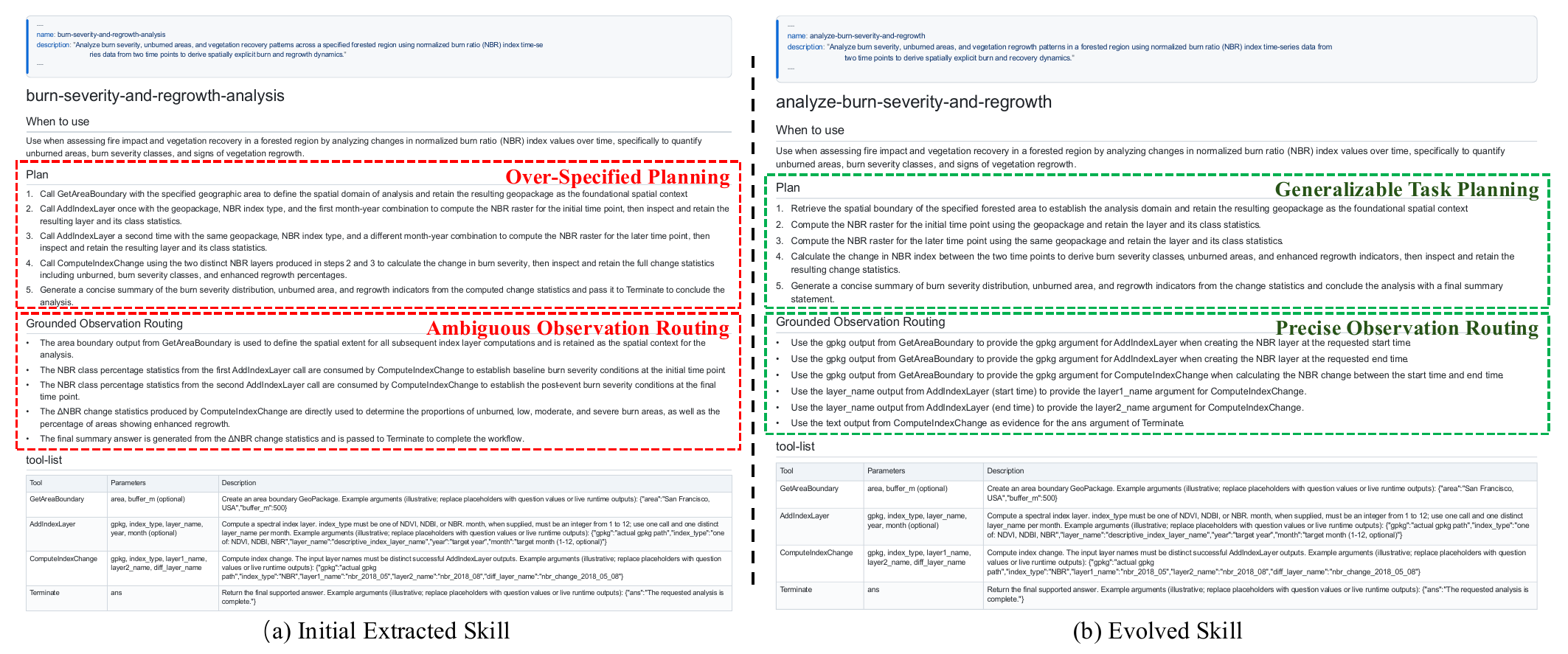}
    \caption{
    Qualitative analysis of a skill before and after evolution.
    (a) The initial extracted skill contains over-specified task planning and ambiguous observation routing.
    (b) The evolved skill adopts generalizable task planning and precise output-to-argument routing.
    }
    \label{fig:skill1}
\end{figure*}

\subsection{Skill Learning and Evolution for Agents}
Skills enable agents to accumulate reusable procedural knowledge.
Early works such as Voyager~\cite{VOYAGER} store acquired skills as executable python code, while ExpeL~\cite{ExpeL} extracts reusable workflows or experience from past tasks.
More recently, Anthropic introduced the Agent Skills standard, representing skills as portable file-based packages~\cite{AgentSkills}.
Building on these foundations, subsequent studies explore automated skill learning and evolution.
EvoDS~\cite{EvoDS} integrates skill learning with context management, MUSE-Autoskill~\cite{MUSEAutoSkill} supports skill lifecycle management, and CoEvoSkills~\cite{CoEvoSkills} employs co-evolutionary verification.
Anything2Skill~\cite{Anything2Skill} constructs skills from heterogeneous knowledge, AutoSkill~\cite{AutoSkill} maintains skills from dialogues, and SkillX~\cite{SkillX}, Trace2Skill~\cite{Trace2Skill}, and SkillRevise~\cite{SkillRevise} refine skills through execution feedback.
Despite these advances, existing methods often treat skills as monolithic text artifacts and rely on outcome-based feedback, providing limited insight into inter-tool dependencies in complex RS workflows.
In contrast, \name grounds skill evolution in actual tool-chain observations, combining dual-judge feedback with a GOR graph to explicitly capture explicit inter-tool dependencies and enable component-level refinement.

%% file: sec/3_method.tex
\section{Method}
\subsection{Component-wise Skill Extraction}
Existing skill extraction methods~\cite{Anything2Skill,AutoSkill,SkillX,Trace2Skill,SkillRevise} usually represent a skill as a monolithic artifact, making refinement coarse-grained and difficult to control: correcting one error may inadvertently corrupt previously correct content.
As shown in Fig.~\ref{fig:overview}(a), to enable controllable evolution, we decompose a skill into independently editable components:
\begin{equation}
s=\bigl(n,d,w,p,r,\ell\bigr),
\end{equation}
where $n$ and $d$ denote the skill name and description, $w$ specifies the activation condition, $p$ defines the execution plan, $r$ captures observation-to-argument dependencies, and $\ell$ records the required tools and their arguments.
Each component is maintained as a Python object and written into the skill file only in the final rendering stage.
Given a sample
\begin{equation}
\xi_i=(q_i,\tau_i,y_i,c_i),\quad
\tau_i=\{(a_{i,t},o_{i,t})\}_{t=1}^{T_i},
\end{equation}
where $q_i$, $y_i$, and $c_i$ denote the query, reference answer, and task type, respectively.
$\tau_i$ represents the tool-call trajectory consisting of $T_i$ sequential pairs of tool invocations $a_{i,t}$ and their corresponding observations $o_{i,t}$.
Each tool invocation is represented as $a_{i,t}=(u_{i,t},p_{i,t})$, where $u_{i,t}$ denotes the invoked tool name and $p_{i,t}$ denotes its corresponding parameters. 
We first initialize a skill template that contains all editable components:
\begin{equation}
\label{deterministic}
s_i^{(0)}
=\bigl(n_i^{(0)},d_i^{(0)},w_i^{(0)},p_i^{(0)},r_i^{(0)},\ell_i\bigr).
\end{equation}
Here, $\ell_i$ is deterministically extracted from the tool list and fixed during refinement, while the remaining components are editable.
The template provides the skill structure and an \emph{Extractor} then independently infers each editable component:
\begin{equation}
\label{extraction}
\hat{s}^{(1)}_i=
\pi_{\theta}(s_i^{(0)},\tau_i;\textbf{P}_{\textrm{ext.}})
=\{n_i^{(1)},d_i^{(1)},w_i^{(1)},p_i^{(1)},r_i^{(1)}\},
\end{equation}
where $\pi_\theta$ is a frozen general-purpose LLM and $\mathbf{P}_{\mathrm{ext.}}$ is the extraction prompt. 
The extracted components are then written back into the template:
\begin{equation}
\label{Overwrite}
s_i^{(1)} = \textrm{Overwrite}_{\mathcal{F}} \left(s_i^{(0)},\hat{s}^{(1)}_i\right), \mathcal{F}=\{n,d,w,p,r\},
\end{equation}
where $\mathcal{F}$ denotes the set of editable components.
The resulting $s_i^{(1)}$ serves as the initialization for subsequent skill evolution. Unlike direct trajectory-to-skill extraction, this template-and-overwrite design combines deterministic structure with LLM-based semantic abstraction, enabling more reliable and controllable skill extraction.

\begin{figure*}[t]
    \centering
    \includegraphics[width=\textwidth]{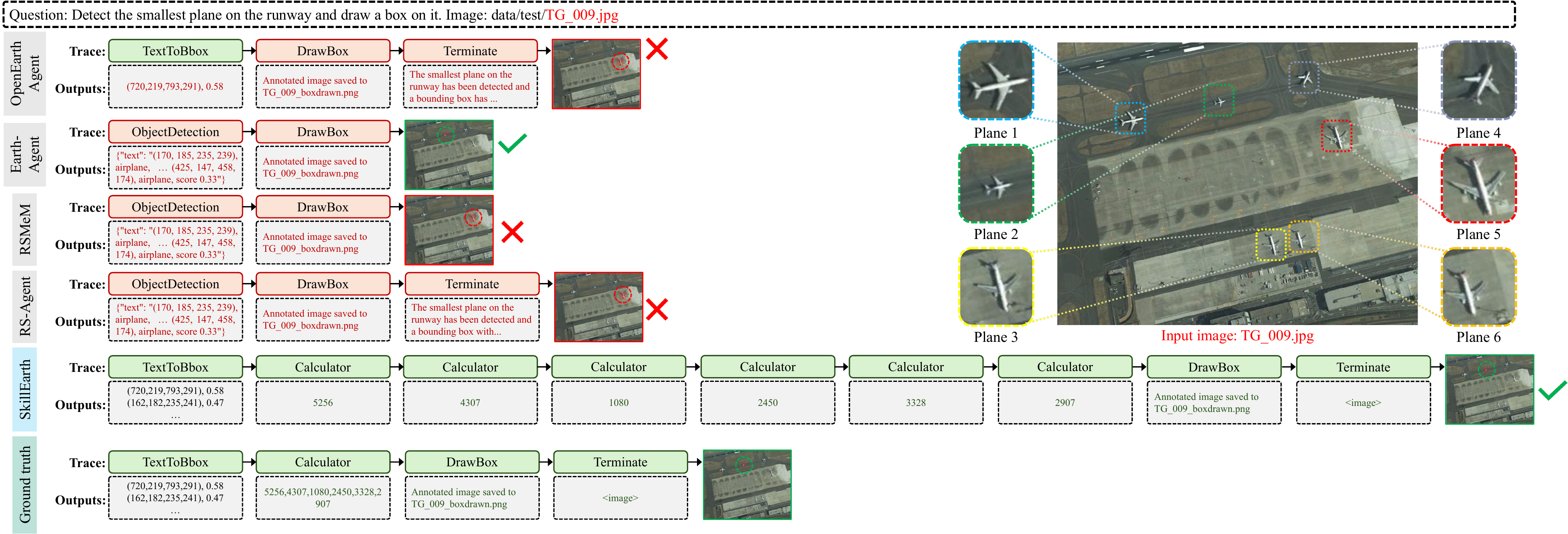}
    \caption{
    Qualitative comparison of \name with state-of-the-art RS agents. 
    We visualize tool-calling trajectories, tool outputs, and final detection results for the smallest airplane. 
    OpenEarthAgent, RSMeM, and RS-Agent fail, while Earth-Agent achieves the correct result via an incorrect tool-calling trajectory. 
    In contrast, \name correctly identifies the smallest airplane by calculating the areas of all six candidate boxes using six \texttt{Calculator} calls (see \textbf{Supp.}~\ref{sec:qual_rs} for detailed analysis).
    }
    \label{fig:comparison_rs}
\end{figure*}

\subsection{Observation-Grounded Skill Evolution}
\label{sec:ogse}
The initial skill $s_i^{(1)}$ provides a reusable solution hypothesis, but its planning may be overly specific and its tool-output dependency descriptions may be unclear, as illustrated in Fig.~\ref{fig:skill1}(a).
We therefore evolve each skill through observation-grounded execution.
The \emph{Evolver} separately verifies (i) trajectory validity, (ii) final-answer semantics, and (iii) inter-tool dependencies, grounding each revision in execution evidence rather than outcome-only reflection.

As shown in Fig.~\ref{fig:overview}(b), given $s_i^{(1)}$ and query $q_i$, an \emph{executor} $\pi_{\theta}(\cdot;\mathbf{P}_{\mathrm{exe.}})$ generates one action at a time, where $\mathbf{P}_{\mathrm{exe.}}$ is execution system prompt.
We use $g\in\{1,\ldots,n_s\}$ for semantic-judge rounds and $k\in\{1,\ldots,n_t\}$ for tool-judge rounds within each semantic round.
At step $t$, the \emph{executor} predicts a tool call conditioned on the query, skill, accepted execution history, and accumulated repair feedback:
\begin{equation}
\label{execution}
\hat{a}_{i,t}^{(g,k)} =
\pi_{\theta}\!\left(
q_i, s_i^{(1)}, \hat{\tau}_{i,<t}^{(g,k)}, \Phi_i^{(g,k)};
\mathbf{P}_{\mathrm{exe.}}
\right),
\end{equation}
where $\hat{\tau}_{i,<t}^{(g,k)}$ denotes the successfully executed trajectory before step $t$, and $\Phi_i^{(g,k)}$ denotes accumulated validation feedback.
This process yields a candidate trajectory $\hat{\tau}_i^{(g,k)}$ and, when applicable, a final answer $\hat{y}_i^{(g,k)}$.

\textbf{Trajectory-level tool validation.}
We first validate each tool-call using a \emph{tool judge} $\pi_{\theta_c}^{c}(\cdot;\mathbf{P}_{\mathrm{tool.}})$ initialized with the system prompt $\mathbf{P}_{\mathrm{tool.}}$.
Rather than evaluating skill text, it checks whether the required tools, execution order, and invocation multiplicity are valid given query $q_i$, reference trajectory $\tau_i$, candidate trajectory $\hat{\tau}_i^{(g,k)}$, and registered tool contracts $\mathcal{C}_i$:
\begin{equation}
\label{tool_judge}
\left(b_{i,g,k}^{\mathrm{tool}}, \phi_{i,g,k}^{\mathrm{tool}}\right) =
\pi_{\theta_c}^{c}\!\left(
q_i, \tau_i, \hat{\tau}_i^{(g,k)}, \mathcal{C}_i;
\mathbf{P}_{\mathrm{tool.}}
\right),
\end{equation}
where $b_{i,g,k}^{\mathrm{tool}}\in\{0,1\}$ is the tool-sequence verdict, and $\phi_{i,g,k}^{\mathrm{tool}}$ is the repair guidance. 
If the verdict is $0$, the guidance is added to $\Phi_i^{(g,k+1)}$ and the agent regenerates the complete trajectory.
Otherwise, the trajectory proceeds to semantic validation.
Notably, each semantic round allows at most $n_t$ tool-validation attempts.
If all attempts fail, the current sample is discarded.

\textbf{Skill-aware semantic validation.}
A valid tool trajectory does not necessarily produce a correct answer.
We therefore use a \emph{semantic judge} to evaluate the generated answer:
\begin{equation}
\label{semantic_judge}
\left(s_{i,g}^{\mathrm{sem}}, \phi_{i,g}^{\mathrm{sem}}\right) =
\pi_{\theta_c}^{c}\!\left(
q_i, \tau_i, y_i, \hat{\tau}_i^{(g,k_g)}, \hat{y}_i^{(g,k_g)}, s_i^{(1)};
\mathbf{P}_{\mathrm{sem.}}
\right),
\end{equation}
where $\mathbf{P}_{\mathrm{sem.}}$ is the semantic system prompt, $y_i$ is the reference answer, $s_{i,g}^{\mathrm{sem}}$ is the semantic score, and $\phi_{i,g}^{\mathrm{sem}}$ is the repair guidance.
Here, $k_g$ denotes the first tool-validation attempt that passes in round $g$.
For textual outputs, the candidate is accepted when $s_{i,g}^{\mathrm{sem}}>\gamma$, where $\gamma$ is the acceptance threshold.
For image outputs, it must produce the required artifact.
Otherwise, $\phi_{i,g}^{\mathrm{sem}}$ is appended to $\Phi_i^{(g+1,1)}$, triggering a new semantic round.
This process runs for at most $n_s$ semantic rounds, and the current sample is discarded if none of them passes validation.
The accepted trajectory is denoted as $\hat{\tau}_i^\star=\hat{\tau}_i^{(g^\star,k_{g^\star})}$, where $g^\star$ is the first passing semantic round and $k_{g^\star}$ is the corresponding tool round.

\begin{table*}[t]
    \centering
    \caption{Comparison with state-of-the-art RS agents using the Qwen3-4B-Instruct on OpenEarth-Bench. EP and FT denote expert priors and fine-tuning, respectively. End-to-End metrics include F$_1$ scores (Per., Op., Logic., GIS), tool order (AnyOr., SameO., Uni.), and accuracy (Ans., Gen.). The top three results are highlighted in \textcolor{red}{red}, \textcolor{blue}{blue}, and \textcolor{orange}{orange}, respectively.}
    \label{tab:rs_comparison}
    {
    \footnotesize
    \setlength{\tabcolsep}{0.6pt}
    \setlength{\fboxsep}{0.6pt}
    \renewcommand{\arraystretch}{0.95}
    \begin{tabular}{l c c *{14}{c}}
        \toprule
        \multirow{2}{*}{\textbf{Method}}
        & \multirow{2}{*}{\textbf{EP}}
        & \multirow{2}{*}{\textbf{FT}}
        & \multicolumn{5}{c}{\textbf{Step-by-Step}}
        & \multicolumn{9}{c}{\textbf{End-to-End}} \\
        \cmidrule(lr){4-8}
        \cmidrule(lr){9-17}

        & & 
        & \textbf{Ins. $\uparrow$}
        & \textbf{Tool. $\uparrow$}
        & \textbf{ArgN. $\uparrow$}
        & \textbf{ArgV. $\uparrow$}
        & \textbf{Summ. $\uparrow$}
        & \textbf{Per. $\uparrow$}
        & \textbf{Op. $\uparrow$}
        & \textbf{Logic. $\uparrow$}
        & \textbf{GIS. $\uparrow$}
        & \textbf{AnyOr. $\uparrow$}
        & \textbf{SameO. $\uparrow$}
        & \textbf{Uni. $\uparrow$}
        & \textbf{Ans. $\uparrow$}
        & \textbf{Gen. $\uparrow$} \\
        \midrule

        OpenEarthAgent \venue{ECCV'26}{OpenEarthAgentSFT}
        & \ding{55}
        & \ding{55}
        & 97.34
        & 86.94
        & 84.12
        & \textcolor{blue}{\textbf{33.55}}
        & 83.28
        & 13.86
        & \textcolor{blue}{\textbf{20.95}}
        & \textcolor{orange}{\textbf{18.35}}
        & \textcolor{blue}{\textbf{71.82}}
        & \textcolor{blue}{\textbf{16.00}}
        & \textcolor{blue}{\textbf{14.71}}
        & \textcolor{blue}{\textbf{21.47}}
        & \textcolor{orange}{\textbf{13.72}}
        & 15.86 \\

        Earth-Agent \venue{ICLR'26}{EarthAgentEO}
        & \ding{55}
        & \ding{55}
        & \textcolor{blue}{\textbf{99.98}}
        & \textcolor{blue}{\textbf{90.84}}
        & \textcolor{red}{\textbf{87.31}}
        & \textcolor{orange}{\textbf{33.37}}
        & \textcolor{blue}{\textbf{88.70}}
        & \textcolor{blue}{\textbf{31.26}}
        & 14.67
        & 11.94
        & 51.58
        & 2.48
        & 2.31
        & 4.19
        & 12.82
        & 19.31 \\

        RSMeM \venue{ACL'26}{RSMeM}
        & \ding{51}
        & \ding{55}
        & \textcolor{orange}{\textbf{99.20}}
        & \textcolor{orange}{\textbf{87.97}}
        & \textcolor{orange}{\textbf{85.06}}
        & 33.05
        & 84.92
        & \textcolor{orange}{\textbf{31.20}}
        & 7.78
        & 15.08
        & \textcolor{orange}{\textbf{53.56}}
        & \textcolor{orange}{\textbf{9.41}}
        & \textcolor{orange}{\textbf{8.98}}
        & \textcolor{orange}{\textbf{11.46}}
        & \textcolor{blue}{\textbf{18.19}}
        & \textcolor{blue}{\textbf{33.10}} \\

        RS-Agent \venue{Sci. China Inf. Sci.'26}{RSAgent}
        & \ding{51}
        & \ding{55}
        & \textcolor{red}{\textbf{100.00}}
        & 43.66
        & 40.17
        & 11.04
        & \textcolor{orange}{\textbf{88.31}}
        & 28.52
        & \textcolor{orange}{\textbf{19.67}}
        & \textcolor{blue}{\textbf{18.72}}
        & 12.67
        & 3.93
        & 3.93
        & 5.82
        & 9.46
        & \textcolor{red}{\textbf{37.93}} \\

        \rowcolor{blue!8}
        \name
        & \ding{55}
        & \ding{55}
        & 99.07
        & \textcolor{red}{\textbf{91.04}}
        & \textcolor{blue}{\textbf{86.48}}
        & \textcolor{red}{\textbf{36.17}}
        & \textcolor{red}{\textbf{92.67}}
        & \textcolor{red}{\textbf{35.00}}
        & \textcolor{red}{\textbf{33.06}}
        & \textcolor{red}{\textbf{28.73}}
        & \textcolor{red}{\textbf{81.85}}
        & \textcolor{red}{\textbf{25.32}}
        & \textcolor{red}{\textbf{25.15}}
        & \textcolor{red}{\textbf{27.89}}
        & \textcolor{red}{\textbf{19.59}}
        & \textcolor{orange}{\textbf{24.14}} \\

        \bottomrule
    \end{tabular}
    }
\end{table*}

\begin{table*}[t]
    \centering
    \caption{Comparison with state-of-the-art RS agents with the Qwen3-4B-Instruct on Earth-Bench under AP and IF settings. The top three results are highlighted in \textcolor{red}{red}, \textcolor{blue}{blue}, and \textcolor{orange}{orange}, respectively.}
    \label{tab:EB}
    {
    \footnotesize
    \setlength{\tabcolsep}{0.7pt}
    \renewcommand{\arraystretch}{0.95}
    \setlength{\fboxsep}{0.5pt}
    \begin{tabular}{@{}l c c *{12}{c}@{}}
        \toprule
        \multirow{3}{*}{\textbf{Method}}
        & \multirow{3}{*}{\textbf{EP}}
        & \multirow{3}{*}{\textbf{FT}}
        & \multicolumn{6}{c}{\textbf{Automatic Planning (AP)}}
        & \multicolumn{6}{c}{\textbf{Instruction Following (IF)}} \\
        \cmidrule(lr){4-9}
        \cmidrule(lr){10-15}
        & &
        & \multicolumn{4}{c}{\textbf{Step-by-Step}}
        & \multicolumn{2}{c}{\textbf{End-to-End}}
        & \multicolumn{4}{c}{\textbf{Step-by-Step}}
        & \multicolumn{2}{c}{\textbf{End-to-End}} \\
        \cmidrule(lr){4-7}
        \cmidrule(lr){8-9}
        \cmidrule(lr){10-13}
        \cmidrule(lr){14-15}
        & &
        & \textbf{AnyOr.$\uparrow$}
        & \textbf{InOr.$\uparrow$}
        & \textbf{Exact.$\uparrow$}
        & \textbf{Param.$\uparrow$}
        & \textbf{Eff.$\downarrow$}
        & \textbf{Acc.$\uparrow$}
        & \textbf{AnyOr.$\uparrow$}
        & \textbf{InOr.$\uparrow$}
        & \textbf{Exact.$\uparrow$}
        & \textbf{Param.$\uparrow$}
        & \textbf{Eff.$\downarrow$}
        & \textbf{Acc.$\uparrow$} \\
        \midrule

        OpenEarthAgent \venue{ECCV'26}{OpenEarthAgentSFT}
        & \ding{55} & \ding{55}
        & 46.55
        & \textcolor{orange}{\textbf{39.01}}
        & \textcolor{orange}{\textbf{28.59}}
        & \textcolor{orange}{\textbf{18.98}}
        & 2.7099
        & 18.15
        & \textcolor{orange}{\textbf{50.21}}
        & \textcolor{orange}{\textbf{41.68}}
        & \textcolor{orange}{\textbf{30.75}}
        & \textcolor{orange}{\textbf{20.75}}
        & 2.7778
        & 19.76 \\

        Earth-Agent \venue{ICLR'26}{EarthAgentEO}
        & \ding{55} & \ding{55}
        & 20.47 & 1.32 & 0.00 & 0.00 & 3.9747 & 15.73
        & 29.69 & 10.89 & 8.63 & 4.21 & 5.4554 & 10.08 \\

        RSMeM \venue{ACL'26}{RSMeM}
        & \ding{51} & \ding{55}
        & \textcolor{orange}{\textbf{50.43}}
        & 33.25 & 7.68 & 7.51
        & \textcolor{red}{\textbf{1.1703}}
        & 21.46
        & 40.72 & 23.27 & 13.93 & 1.16
        & \textcolor{red}{\textbf{0.4839}}
        & 21.46 \\

        HiRS-Agent \venue{ACM MM'26}{HiRS-Agent}
        & \ding{51} & \ding{51}
        & \textcolor{blue}{\textbf{50.95}}
        & \textcolor{blue}{\textbf{41.10}}
        & \textcolor{blue}{\textbf{31.67}}
        & \textcolor{blue}{\textbf{19.54}}
        & 2.5059
        & \textcolor{red}{\textbf{43.95}}
        & \textcolor{blue}{\textbf{53.79}}
        & \textcolor{blue}{\textbf{46.34}}
        & \textcolor{blue}{\textbf{34.64}}
        & \textcolor{blue}{\textbf{20.81}}
        & 3.3013
        & \textcolor{red}{\textbf{45.56}} \\

        RS-Agent \venue{Sci. China Inf. Sci.'26}{RSAgent}
        & \ding{51} & \ding{55}
        & 22.68 & 15.84 & 15.17 & 8.90
        & \textcolor{orange}{\textbf{2.3385}}
        & \textcolor{blue}{\textbf{29.44}}
        & 27.25 & 18.61 & 17.74 & 9.18
        & \textcolor{orange}{\textbf{2.2295}}
        & \textcolor{blue}{\textbf{33.06}} \\

        \rowcolor{blue!8}
        SkillEarth
        & \ding{55} & \ding{55}
        & \textcolor{red}{\textbf{53.17}}
        & \textcolor{red}{\textbf{43.24}}
        & \textcolor{red}{\textbf{35.42}}
        & \textcolor{red}{\textbf{22.54}}
        & \textcolor{blue}{\textbf{1.8659}}
        & \textcolor{orange}{\textbf{22.98}}
        & \textcolor{red}{\textbf{60.18}}
        & \textcolor{red}{\textbf{51.38}}
        & \textcolor{red}{\textbf{47.64}}
        & \textcolor{red}{\textbf{28.24}}
        & \textcolor{blue}{\textbf{1.9647}}
        & \textcolor{orange}{\textbf{26.61}} \\

        \bottomrule
    \end{tabular}
    }
\end{table*}

\textbf{Evidence-grounded skill refinement.}
Once the execution is validated, the accepted trajectory is used to refine the skill components:
\begin{equation}
\label{refine1}
\left(
n_i^{(2)}, d_i^{(2)},w_i^{(2)},p_i^{(2)}
\right)
=
\pi_{\theta}\!\left(
q_i, \hat{\tau}_i^\star, s_i^{(1)};
\mathbf{P}_{\mathrm{ref.}}
\right).
\end{equation}
Here, $\mathbf{P}_{\mathrm{ref.}}$ denotes the skill-refinement system prompt.
To further refine inter-tool dependencies, we construct a GOR graph $\mathcal{G}_i=(\mathcal{V}_i,\mathcal{E}_i)$ from the accepted trajectory, where each node represents a successful tool invocation:
\begin{equation}
v_{i,t}=
\left(
u_{i,t},p_{i,t},\mathcal{Z}_{i,t}
\right)
\in\mathcal{V}_i,
\end{equation}
where $u_{i,t}$, $p_{i,t}$ and $\mathcal{Z}_{i,t}$ denote the $t$-th tool name, arguments, and execution result.
For the $k$-th intermediate result $z_{i,m,k}$ generated by the $m$-th invocation and consumed by the $j$-th argument $p_{i,n,j}$ of the $n$-th invocation, we add a routing edge:
\begin{equation}
e_{i,m,k\rightarrow i,n,j}
=
\left(
v_{i,m},z_{i,m,k},v_{i,n},p_{i,n,j}
\right)
\in\mathcal{E}_i,
\quad m<n.
\end{equation}
The GOR graph thus captures tool-call order and observation routing between dependent tool calls.
We then translate it into natural-language routing rules using $\mathbf{P}_{\mathrm{GOR.}}$:
\begin{equation}
\label{refine2}
r_i^{(2)}
=
\pi_{\theta}\left(
\mathcal{G}_i;\mathbf{P}_{\mathrm{GOR.}}
\right).
\end{equation}
Finally, the evolved skill is obtained as:
\begin{equation}
s_i^{(2)}=
\bigl(
n_i^{(2)},d_i^{(2)},w_i^{(2)},p_i^{(2)},r_i^{(2)},\ell_i
\bigr).
\end{equation}
Through three independent validation stages, the \emph{Evolver} enables fine-grained, controllable skill evolution based on explicit execution evidence rather than coarse-grained outcome reflection.
Fig.~\ref{fig:skill1}(b) illustrates the evolved skill, featuring more generalizable task planning and precise observation routing (See \textbf{Supp.}~\ref{sec:skill_evolution_case} for detailed analysis).

\begin{table*}[t]
    \centering
    \caption{Comparison with skill extraction methods using the Qwen3-4B-Instruct on OpenEarth-Bench. End-to-End metrics include F$_1$ scores (Per., Op., Logic., GIS), tool order (AnyOr., SameO., Uni.), and accuracy (Ans., Gen.). The top three results are highlighted in \textcolor{red}{red}, \textcolor{blue}{blue}, and \textcolor{orange}{orange}, respectively.}
    \label{tab:skill_comparison}
    {
    \footnotesize
    \setlength{\tabcolsep}{0.6pt}
    \setlength{\fboxsep}{0.6pt}
    \renewcommand{\arraystretch}{0.95}
    \begin{tabular}{l *{14}{c}}
        \toprule
        \multirow{2}{*}{\textbf{Method}}
        & \multicolumn{5}{c}{\textbf{Step-by-Step}}
        & \multicolumn{9}{c}{\textbf{End-to-End}} \\
        \cmidrule(lr){2-6}
        \cmidrule(lr){7-15}

        & \textbf{Ins. $\uparrow$}
        & \textbf{Tool. $\uparrow$}
        & \textbf{ArgN. $\uparrow$}
        & \textbf{ArgV. $\uparrow$}
        & \textbf{Summ. $\uparrow$}
        & \textbf{Per. $\uparrow$}
        & \textbf{Op. $\uparrow$}
        & \textbf{Logic. $\uparrow$}
        & \textbf{GIS. $\uparrow$}
        & \textbf{AnyOr. $\uparrow$}
        & \textbf{SameO. $\uparrow$}
        & \textbf{Uni. $\uparrow$}
        & \textbf{Ans. $\uparrow$}
        & \textbf{Gen. $\uparrow$} \\
        \midrule

        Trace2Skill \venue{arXiv'26}{Trace2Skill}
        & \textcolor{orange}{\textbf{98.88}}
        & \textcolor{orange}{\textbf{89.89}}
        & \textcolor{orange}{\textbf{85.61}}
        & \textcolor{blue}{\textbf{36.35}}
        & 84.43
        & 13.21
        & \textcolor{blue}{\textbf{30.56}}
        & \textcolor{blue}{\textbf{22.97}}
        & \textcolor{orange}{\textbf{40.33}}
        & 2.40
        & 2.40
        & 3.51
        & \textcolor{orange}{\textbf{9.67}}
        & \textcolor{orange}{\textbf{14.48}} \\

        AutoSkill \venue{arXiv'26}{AutoSkill}
        & 98.78
        & 83.60
        & 79.61
        & \textcolor{red}{\textbf{38.35}}
        & \textcolor{blue}{\textbf{87.31}}
        & \textcolor{blue}{\textbf{17.97}}
        & 24.88
        & 20.02
        & 35.96
        & \textcolor{orange}{\textbf{3.85}}
        & \textcolor{orange}{\textbf{3.76}}
        & \textcolor{orange}{\textbf{4.96}}
        & 8.31
        & 9.66 \\

        SkillX \venue{arXiv'26}{SkillX}
        & \textcolor{blue}{\textbf{99.03}}
        & \textcolor{blue}{\textbf{90.37}}
        & \textcolor{blue}{\textbf{85.62}}
        & 34.19
        & \textcolor{orange}{\textbf{84.68}}
        & \textcolor{orange}{\textbf{15.80}}
        & \textcolor{orange}{\textbf{29.72}}
        & 20.12
        & \textcolor{blue}{\textbf{72.62}}
        & \textcolor{blue}{\textbf{17.11}}
        & \textcolor{blue}{\textbf{16.77}}
        & \textcolor{blue}{\textbf{20.87}}
        & \textcolor{red}{\textbf{20.12}}
        & \textcolor{blue}{\textbf{17.24}} \\

        SkillRevise \venue{EMNLP'26}{SkillRevise}
        & 98.51
        & 88.30
        & 84.07
        & 33.60
        & 83.83
        & 13.50
        & 25.26
        & \textcolor{orange}{\textbf{21.31}}
        & 35.50
        & 2.40
        & 2.22
        & 3.25
        & 6.63
        & \textcolor{orange}{\textbf{14.48}} \\

        \rowcolor{blue!8}
        \name
        & \textcolor{red}{\textbf{99.07}}
        & \textcolor{red}{\textbf{91.04}}
        & \textcolor{red}{\textbf{86.48}}
        & \textcolor{orange}{\textbf{36.17}}
        & \textcolor{red}{\textbf{92.67}}
        & \textcolor{red}{\textbf{35.00}}
        & \textcolor{red}{\textbf{33.06}}
        & \textcolor{red}{\textbf{28.73}}
        & \textcolor{red}{\textbf{81.85}}
        & \textcolor{red}{\textbf{25.32}}
        & \textcolor{red}{\textbf{25.15}}
        & \textcolor{red}{\textbf{27.89}}
        & \textcolor{blue}{\textbf{19.59}}
        & \textcolor{red}{\textbf{24.14}} \\

        \bottomrule
    \end{tabular}
    }
\end{table*}

\begin{table*}[t]
    \centering
    \caption{Comparison with skill extraction methods using the Qwen3-4B-Instruct on Earth-Bench under AP and IF settings. The top three results are highlighted in \textcolor{red}{red}, \textcolor{blue}{blue}, and \textcolor{orange}{orange}, respectively.}
    \label{tab:skill_EB}
    {
    \footnotesize
    \setlength{\tabcolsep}{0.7pt}
    \renewcommand{\arraystretch}{0.95}
    \setlength{\fboxsep}{0.5pt}
    \begin{tabular}{@{}l *{12}{c}@{}}
        \toprule
        \multirow{3}{*}{\textbf{Method}}
        & \multicolumn{6}{c}{\textbf{Automatic Planning (AP)}}
        & \multicolumn{6}{c}{\textbf{Instruction Following (IF)}} \\
        \cmidrule(lr){2-7}
        \cmidrule(lr){8-13}
        & \multicolumn{4}{c}{\textbf{Step-by-Step}}
        & \multicolumn{2}{c}{\textbf{End-to-End}}
        & \multicolumn{4}{c}{\textbf{Step-by-Step}}
        & \multicolumn{2}{c}{\textbf{End-to-End}} \\
        \cmidrule(lr){2-5}
        \cmidrule(lr){6-7}
        \cmidrule(lr){8-11}
        \cmidrule(lr){12-13}
        & \textbf{AnyOr.$\uparrow$}
        & \textbf{InOr.$\uparrow$}
        & \textbf{Exact.$\uparrow$}
        & \textbf{Param.$\uparrow$}
        & \textbf{Eff.$\downarrow$}
        & \textbf{Acc.$\uparrow$}
        & \textbf{AnyOr.$\uparrow$}
        & \textbf{InOr.$\uparrow$}
        & \textbf{Exact.$\uparrow$}
        & \textbf{Param.$\uparrow$}
        & \textbf{Eff.$\downarrow$}
        & \textbf{Acc.$\uparrow$} \\
        \midrule

        Trace2Skill \venue{arXiv'26}{Trace2Skill}
        & 46.14
        & 33.03
        & \textcolor{orange}{\textbf{26.34}}
        & \textcolor{orange}{\textbf{20.48}}
        & 1.8854
        & \textcolor{orange}{\textbf{20.81}}
        & \textcolor{orange}{\textbf{49.55}}
        & \textcolor{orange}{\textbf{39.39}}
        & \textcolor{orange}{\textbf{32.53}}
        & \textcolor{orange}{\textbf{24.29}}
        & 2.0532
        & \textcolor{blue}{\textbf{24.60}} \\

        AutoSkill \venue{arXiv'26}{AutoSkill}
        & 43.56
        & 28.63
        & 17.21
        & 13.12
        & \textcolor{red}{\textbf{1.6711}}
        & 18.55
        & 48.34
        & 35.44
        & 25.52
        & 18.01
        & \textcolor{blue}{\textbf{1.8344}}
        & 22.18 \\

        SkillX \venue{arXiv'26}{SkillX}
        & \textcolor{blue}{\textbf{51.54}}
        & \textcolor{blue}{\textbf{42.95}}
        & \textcolor{blue}{\textbf{33.43}}
        & \textcolor{blue}{\textbf{20.66}}
        & 1.9922
        & 20.19
        & \textcolor{blue}{\textbf{58.00}}
        & \textcolor{blue}{\textbf{48.87}}
        & \textcolor{blue}{\textbf{44.17}}
        & \textcolor{blue}{\textbf{27.27}}
        & 1.9772
        & 18.15 \\

        SkillRevise \venue{EMNLP'26}{SkillRevise}
        & \textcolor{orange}{\textbf{46.42}}
        & \textcolor{orange}{\textbf{34.68}}
        & 24.93
        & 19.22
        & \textcolor{blue}{\textbf{1.7609}}
        & \textcolor{blue}{\textbf{22.18}}
        & 47.65
        & 35.79
        & 30.03
        & 21.86
        & \textcolor{red}{\textbf{1.8034}}
        & \textcolor{orange}{\textbf{22.58}} \\

        \rowcolor{blue!8}
        SkillEarth
        & \textcolor{red}{\textbf{53.17}}
        & \textcolor{red}{\textbf{43.24}}
        & \textcolor{red}{\textbf{35.42}}
        & \textcolor{red}{\textbf{22.54}}
        & \textcolor{orange}{\textbf{1.8659}}
        & \textcolor{red}{\textbf{22.98}}
        & \textcolor{red}{\textbf{60.18}}
        & \textcolor{red}{\textbf{51.38}}
        & \textcolor{red}{\textbf{47.64}}
        & \textcolor{red}{\textbf{28.24}}
        & \textcolor{orange}{\textbf{1.9647}}
        & \textcolor{red}{\textbf{26.61}} \\

        \bottomrule
    \end{tabular}
    }
\end{table*}

\subsection{Skill Decision and Application}
As shown in Fig.~\ref{fig:overview}(c), once the skill has been evolved, the \emph{Decider} determines how to integrate $s_i^{(2)}$ into the skill bank $\mathcal{B}$.
For each $s_j\in\mathcal{B}$, it first performs structural matching by task type and tool-call sequence:
\begin{equation}
\label{matching}
C_{ij}=
\mathbb{I}[c_i=c_j]\cdot
\mathbb{I}[\boldsymbol{\sigma}_i=\boldsymbol{\sigma}_j],
\end{equation}
where $c_i$ and $c_j$ denote task types, and $\boldsymbol{\sigma}_i$ and $\boldsymbol{\sigma}_j$ denote tool-call sequences.
If $c_i\neq c_j$ or $\boldsymbol{\sigma}_i\neq\boldsymbol{\sigma}_j$, $s_i^{(2)}$ is directly added to $\mathcal{B}$ as a new task type or workflow.
Only when $C_{ij}=1$, the \emph{Decider} determines the integration decision:
\begin{equation}
\label{decision}
d_{ij} = \pi_{\theta}(s_i^{(2)},s_j;\mathbf{P}_{\mathrm{dec.}}),
\end{equation}
where $d_{ij}\in\{\mathrm{Add},\mathrm{Merge},\mathrm{Discard}\}$.
$\mathrm{Add}$ inserts $s_i^{(2)}$ into $\mathcal{B}$, $\mathrm{Discard}$ removes it, and $\mathrm{Merge}$ combines it with the most semantically similar skill $s_j^{*}$.
For merging, an LLM-based merger with prompt $\mathbf{P}_{\mathrm{mer.}}$ generates:
\begin{equation}
\label{merging}
s_{ij}^{*}
=
\pi_{\theta}
(s_i^{(2)},s_j^*;\mathbf{P}_{\mathrm{mer.}}),
\end{equation}
where $s_{ij}^{*}$ denotes the merged skill.
The detailed pseudocode for SkillBank construction is provided in \textbf{Supp.}~\ref{sec:algorithm}.

During application, the agent retrieves the most relevant skill $s^*$ from $\mathcal{B}$ via cosine similarity and incorporates it into $\mathbf{P}_{\mathrm{exe.}}$ (see Fig.~\ref{fig:overview}(d)).
At step $t$, the agent predicts:
\begin{equation}
a_t
=
\pi_{\theta}
(q,s^{*},\tau_{<t};\mathbf{P}_{\mathrm{exe.}}),
\end{equation}
where $\tau_{<t}$ denotes the trajectory history before step $t$.
The corresponding tool executes $a_t$ to obtain $o_t$, repeating until a final answer is generated.
By storing RS knowledge in skill files, $\mathcal{B}$ enables cross-LLM reuse without retraining.
Fig.~\ref{fig:comparison_rs} presents a qualitative comparison with existing RS agents, highlighting the effectiveness of skill-guided tool planning and execution (see \textbf{Supp.}~\ref{sec:qual_rs} for detailed analysis).

%% file: sec/4_experiments.tex
\begin{table*}[t]
    \centering
    \caption{
    Skill generalizability across different LLMs on OpenEarth-Bench. All models use a maximum output token budget of 1024. End-to-End metrics include F$_1$ scores (Per., Op., Logic., GIS), tool order (AnyOr., SameO., Uni.), and accuracy (Ans., Gen.).
    \textbf{bold} indicates the best result for each model.
    For each metric, the top three models are highlighted with \colorbox{red!12}{red}, \colorbox{cyan!12}{cyan}, and \colorbox{green!12}{green} backgrounds, respectively.
    }
    \label{tab:general_comparison}
    {
    \footnotesize
    \setlength{\tabcolsep}{0.6pt}
    \setlength{\fboxsep}{0.5pt}
    \renewcommand{\arraystretch}{0.70}
    \resizebox{\textwidth}{!}{%
    \begin{tabular}{l l c *{14}{c}}
        \toprule
        \multirow{2}{*}{\textbf{Model}}
        & \multirow{2}{*}{\textbf{Setting}}
        & \multirow{2}{*}{\textbf{Param.}}
        & \multicolumn{5}{c}{\textbf{Step-by-Step}}
        & \multicolumn{9}{c}{\textbf{End-to-End}} \\
        \cmidrule(lr){4-8}
        \cmidrule(lr){9-17}

        & &
        & \textbf{Inst. $\uparrow$}
        & \textbf{Tool. $\uparrow$}
        & \textbf{ArgN. $\uparrow$}
        & \textbf{ArgV. $\uparrow$}
        & \textbf{Summ. $\uparrow$}
        & \textbf{Per. $\uparrow$}
        & \textbf{Op. $\uparrow$}
        & \textbf{Logic. $\uparrow$}
        & \textbf{GIS. $\uparrow$}
        & \textbf{AnyOr. $\uparrow$}
        & \textbf{SameO. $\uparrow$}
        & \textbf{Uni. $\uparrow$}
        & \textbf{Ans. $\uparrow$}
        & \textbf{Gen. $\uparrow$} \\
        \midrule

        \rowcolor{blue!8}
        \multicolumn{17}{c}{\textsc{Frontier LLMs}} \\
        \midrule

        \multirow{2}{*}{GPT-5.6-Terra~\cite{openai2026gpt56}}
        & w/o skill & --
        & 96.35
        & 81.88
        & 79.34
        & 37.95
        & \colorbox{cyan!12}{96.88}
        & 30.99
        & \rankthree{46.40}
        & 6.76
        & \ranktwo{92.63}
        & 45.17
        & 45.00
        & 47.13
        & 50.82
        & 35.86 \\

        & \textbf{w/ skill} & --
        & \textbf{98.46}
        & \textbf{88.35}
        & \colorbox{green!12}{\textbf{84.99}}
        & \colorbox{cyan!12}{\textbf{43.07}}
        & \colorbox{red!12}{\textbf{97.30}}
        & \rankone{\textbf{44.37}}
        & \rankone{\textbf{65.59}}
        & \textbf{20.50}
        & \rankone{\textbf{95.59}}
        & \ranktwo{\textbf{51.84}}
        & \ranktwo{\textbf{51.33}}
        & \rankthree{\textbf{55.18}}
        & \ranktwo{\textbf{53.91}}
        & \textbf{51.72} \\
        \cmidrule(lr){2-17}

        \multirow{2}{*}{Doubao-Seed-2.0-Mini~\cite{bytedance2026seed2}}
        & w/o skill & --
        & \colorbox{green!12}{\textbf{99.83}}
        & 88.72
        & 82.46
        & 36.12
        & 94.02
        & \textbf{34.74}
        & \textbf{28.19}
        & 19.49
        & 87.18
        & 33.96
        & 33.62
        & 38.84
        & 36.63
        & \ranktwo{\textbf{56.55}} \\

        & \textbf{w/ skill} & --
        & \colorbox{cyan!12}{\textbf{99.86}}
        & \colorbox{cyan!12}{\textbf{89.75}}
        & \colorbox{red!12}{\textbf{86.50}}
        & \textbf{39.46}
        & \textbf{96.13}
        & \ranktwo{\textbf{43.58}}
        & 20.69
        & \textbf{27.23}
        & \rankthree{\textbf{89.09}}
        & \textbf{45.51}
        & \textbf{44.82}
        & \textbf{49.70}
        & \textbf{43.78}
        & 51.03 \\
        \cmidrule(lr){2-17}

        \multirow{2}{*}{DeepSeek-V4-Flash~\cite{deepseek2026v4}}
        & w/o skill & 284B
        & 99.78
        & 84.92
        & 78.90
        & \colorbox{green!12}{40.32}
        & 96.64
        & 26.05
        & 38.51
        & 13.11
        & 82.27
        & \rankthree{50.90}
        & \rankthree{50.21}
        & \ranktwo{56.54}
        & \rankthree{53.41}
        & \rankthree{55.86} \\

        & \textbf{w/ skill} & 284B
        & \colorbox{red!12}{\textbf{99.92}}
        & \textbf{89.04}
        & \textbf{82.37}
        & \colorbox{red!12}{\textbf{43.75}}
        & \colorbox{green!12}{\textbf{96.69}}
        & \textbf{32.30}
        & \ranktwo{\textbf{51.43}}
        & \ranktwo{\textbf{29.96}}
        & \textbf{82.30}
        & \rankone{\textbf{55.43}}
        & \rankone{\textbf{54.41}}
        & \rankone{\textbf{61.93}}
        & \rankone{\textbf{54.67}}
        & \rankone{\textbf{59.31}} \\

        \midrule
        \rowcolor{blue!8}
        \multicolumn{17}{c}{\textsc{Open-Source LLMs}} \\
        \midrule

        \multirow{2}{*}{Mistral-v0.3-Instruct~\cite{mistralai2024mistral7b}}
        & w/o skill & 7B
        & 60.76
        & 48.87
        & 39.69
        & 20.63
        & \textbf{80.21}
        & 17.86
        & 14.24
        & 7.12
        & 29.16
        & 6.07
        & 5.65
        & 7.61
        & 0.25
        & 6.21 \\

        & \textbf{w/ skill} & 7B
        & \textbf{63.14}
        & \textbf{54.30}
        & \textbf{43.66}
        & \textbf{20.64}
        & 70.70
        & \textbf{28.79}
        & \textbf{19.34}
        & \textbf{27.43}
        & \textbf{48.55}
        & \textbf{18.22}
        & \textbf{17.28}
        & \textbf{23.61}
        & \textbf{2.52}
        & \textbf{13.79} \\
        \cmidrule(lr){2-17}

        \multirow{2}{*}{DeepSeek-R1~\cite{deepseek2025r1}}
        & w/o skill & 7B
        & 44.34
        & 28.33
        & 23.99
        & 7.45
        & \textbf{87.92}
        & 4.43
        & \textbf{12.59}
        & 2.90
        & \textbf{18.97}
        & 4.53
        & 2.74
        & 5.73
        & 0.09
        & \textbf{13.10} \\

        & \textbf{w/ skill} & 7B
        & \textbf{55.95}
        & \textbf{42.90}
        & \textbf{37.12}
        & \textbf{12.88}
        & 78.14
        & \textbf{12.09}
        & 8.53
        & \textbf{13.53}
        & 13.74
        & \textbf{5.82}
        & \textbf{4.96}
        & \textbf{9.24}
        & \textbf{0.92}
        & 8.97 \\
        \cmidrule(lr){2-17}

        \multirow{2}{*}{Qwen2.5-7B-Instruct~\cite{qwen2024qwen25}}
        & w/o skill & 7B
        & 95.34
        & 86.46
        & 79.15
        & 38.03
        & 84.17
        & 25.35
        & 30.69
        & 27.97
        & 67.99
        & 37.72
        & 37.30
        & 40.89
        & 14.49
        & 43.45 \\

        & \textbf{w/ skill} & 7B
        & \textbf{97.63}
        & \colorbox{green!12}{\textbf{89.50}}
        & \textbf{81.97}
        & \textbf{38.24}
        & \textbf{85.60}
        & \rankthree{\textbf{39.20}}
        & \textbf{40.22}
        & \rankone{\textbf{37.24}}
        & \textbf{71.26}
        & \textbf{43.11}
        & \textbf{42.86}
        & \textbf{47.31}
        & \textbf{16.22}
        & \textbf{51.03} \\
        \cmidrule(lr){2-17}

        \multirow{2}{*}{Qwen3-4B-Instruct~\cite{qwen2025qwen3}}
        & w/o skill & 4B
        & 98.47
        & 87.70
        & 84.33
        & 33.67
        & 92.55
        & 13.31
        & 12.54
        & 11.80
        & 76.42
        & 15.23
        & 14.63
        & 18.48
        & 16.94
        & 7.59 \\

        & \textbf{w/ skill} & 4B
        & 99.07
        & \colorbox{red!12}{\textbf{91.04}}
        & \colorbox{cyan!12}{\textbf{86.48}}
        & \textbf{36.17}
        & \textbf{92.67}
        & \textbf{35.00}
        & \textbf{33.06}
        & \rankthree{\textbf{28.73}}
        & \textbf{81.85}
        & \textbf{25.32}
        & \textbf{25.15}
        & \textbf{27.89}
        & \textbf{19.59}
        & \textbf{24.14} \\

        \bottomrule
    \end{tabular}
    }}
\end{table*}

\begin{table}[t]
    \centering
    \caption{Ablation study of SkillEarth. 
    The top three results are highlighted in \textcolor{red}{red}, \textcolor{blue}{blue}, and \textcolor{orange}{orange}, respectively. 
    Complete results are provided in Tabs.~\ref{tab:ablation_e2e} and \ref{tab:ablation_step} of the \textbf{Supp.}~\ref{sec:supp_ablation}.}
    \label{tab:ablation_main}
    \footnotesize
    \setlength{\tabcolsep}{2.5pt}
    \renewcommand{\arraystretch}{0.9}
    \begin{tabular}{lccccc}
        \toprule
        \multirow{2}{*}{\textbf{Setting}}
        & \multicolumn{3}{c}{\textbf{Step-by-Step}}
        & \multicolumn{2}{c}{\textbf{End-to-End}} \\
        \cmidrule(lr){2-4}
        \cmidrule(l){5-6}
        & \textbf{Tool.}
        & \textbf{ArgN.}
        & \textbf{Summ.}
        & \textbf{Ans.}
        & \textbf{Gen.} \\
        \midrule

        Baseline
        & 90.33
        & 85.36
        & 92.01
        & 18.35
        & 18.62 \\

        + Tool Judge
        & \textcolor{blue}{\textbf{90.92}}
        & \textcolor{orange}{\textbf{85.70}}
        & \textcolor{orange}{\textbf{92.51}}
        & \textcolor{orange}{\textbf{18.42}}
        & \textcolor{orange}{\textbf{20.09}} \\

        + Semantic Judge
        & \textcolor{orange}{\textbf{90.81}}
        & \textcolor{blue}{\textbf{85.99}}
        & \textcolor{blue}{\textbf{92.66}}
        & \textcolor{blue}{\textbf{18.49}}
        & \textcolor{blue}{\textbf{23.45}} \\

        \rowcolor{blue!8}
        + GOR Graph (\textit{Ours})
        & \textcolor{red}{\textbf{91.04}}
        & \textcolor{red}{\textbf{86.48}}
        & \textcolor{red}{\textbf{92.67}}
        & \textcolor{red}{\textbf{19.59}}
        & \textcolor{red}{\textbf{24.14}} \\

        \bottomrule
    \end{tabular}
\end{table}

\section{Experiments}
\subsection{Experimental Setup}
\label{sec:4.1}
We evaluate \name on OpenEarth-Bench~\cite{OpenEarthAgentSFT} and Earth-Bench~\cite{EarthAgentEO} using official metrics.
OpenEarth-Bench contains 14,538 training and 1,169 evaluation instances across 60 RS tasks, while Earth-Bench includes 248 expert-curated samples and 13,729 multimodal images under Automatic Planning (AP) and Instruction Following (IF) settings.
Both benchmarks assess tool execution and task outcomes through Step-by-Step and End-to-End evaluations, respectively.
To reduce time cost, we construct the OpenEarth-Bench SkillBank using 1,050 training trajectories (7.2\% of the training set) covering all 60 task types and evaluate on all 1,169 instances (see \textbf{Supp.}~\ref{supp_sec:dataset} for the reason behind this selection).
For Earth-Bench, we evaluate all methods following the official evaluation protocols, without additional processing.
All ablation studies are conducted on OpenEarth-Bench.
We use Doubao-Seed-Evolving~\cite{doubao_seed_evolving} as the cloud-based judge via its publicly accessible API and Qwen3-4B-Instruct~\cite{qwen2025qwen3} as the default backbone.
Experiments run on four NVIDIA A100 GPUs.
Dataset construction, model configurations, and metric definitions are detailed in \textbf{Supp.}~\ref{supp_sec:implementation}, with system prompts in \textbf{Supp.}~\ref{supp_sec:prompts}.

\subsection{Comparison with State-of-the-Art RS Agents}
Tab.~\ref{tab:rs_comparison} compares \name with state-of-the-art RS agents on OpenEarth-Bench.
Without relying on expert priors, \name achieves the best performance on three of five Step-by-Step metrics and eight of nine End-to-End metrics, demonstrating its effectiveness in tool execution and task completion.
Qualitative comparison is provided in Fig.~\ref{fig:comparison_rs}.

Tab.~\ref{tab:EB} reports the results on Earth-Bench.
\name outperforms all methods on each Step-by-Step metric in both AP and IF modes, with the second-best efficiency.
In particular, it improves Exact Match (Exact.) over the strongest method by 3.75\% and 13.00\% in AP and IF modes, respectively.
Although RS-Agent and HiRS-Agent achieve higher End-to-End accuracy, they rely on expert priors, with HiRS-Agent additionally requiring fine-tuning.
These results highlight \name's ability to execute tools effectively and reliably without expert priors or fine-tuning.

\subsection{Comparison with Skill Extraction Methods}
We compare \name with four skill extraction methods using the same Qwen3-4B-Instruct backbone.
As shown in Tab.~\ref{tab:skill_comparison}, \name achieves the best results on four out of five Step-by-Step metrics and eight out of nine End-to-End metrics on OpenEarth-Bench, demonstrating the effectiveness of evolved skills in guiding tool execution.
A qualitative comparison is provided in Fig.~\ref{fig:comparison_skill} of \textbf{Supp.}~\ref{sec:qual_skill}.

Tab.~\ref{tab:skill_EB} shows that \name outperforms all methods across all Step-by-Step metrics on Earth-Bench under both AP and IF settings, with the third-best efficiency.
It also achieves the highest End-to-End accuracy of 22.98\% and 26.61\% under AP and IF, respectively.

\subsection{Generalization and Cross-Model Transfer}
We evaluate the generalization of \name on OpenEarth-Bench using three frontier and four open-source LLMs.
As shown in Tab.~\ref{tab:general_comparison}, skill guidance generally improves both Step-by-Step and End-to-End performance.
However, Mistral-v0.3-Instruct and DeepSeek-R1 show declines in Step-by-Step Summ. accuracy despite improved tool execution.
We attribute this to the fact that while skills can enhance tool execution and task completion, synthesizing tool observations into final answers still relies on the model's own reasoning capabilities.
For End-to-End evaluation, Qwen3-4B-Instruct achieves notable gains in Gen. accuracy.
Performance changes across all nine End-to-End metrics are visualized in Fig.~\ref{fig:bar} of the supplementary material.
Moreover, frontier LLMs generally outperform open-source models, especially in End-to-End performance.

We further evaluate the cross-model transferability of \name.
As shown in Tab.~\ref{tab:transfer_main}, skills extracted by frontier LLMs improve both Step-by-Step and End-to-End performance when transferred to open-source LLMs, and vice versa.
These results demonstrate that \name learns transferable RS knowledge that can be reused across different LLMs without retraining.

\begin{table}[t]
    \centering
    \caption{Cross-model skill transfer on OpenEarth-Bench.
    ``Skill Source'' denotes the model used to extract the transferred skills.
    The best and second-best results within each evaluation model are highlighted in
    \textcolor{red}{red} and \textcolor{blue}{blue}, respectively.
    Complete results are provided in
    Tabs.~\ref{tab:transfer_step} and \ref{tab:transfer_e2e}.}
    \label{tab:transfer_main}
    \footnotesize
    \setlength{\tabcolsep}{2.5pt}
    \renewcommand{\arraystretch}{0.9}
    \resizebox{\linewidth}{!}{
    \begin{tabular}{llccccc}
        \toprule
        \multirow{2}{*}{\textbf{Evaluation Model}}
        & \multirow{2}{*}{\textbf{Skill Source}}
        & \multicolumn{3}{c}{\textbf{Step-by-Step}}
        & \multicolumn{2}{c}{\textbf{End-to-End}} \\
        \cmidrule(lr){3-5}
        \cmidrule(l){6-7}
        & & \textbf{Tool.}
        & \textbf{ArgN.}
        & \textbf{Summ.}
        & \textbf{Ans.}
        & \textbf{Gen.} \\
        \midrule

        \multirow{4}{*}{Qwen3-4B-Instruct}
        & --
        & 87.70 & 84.33 & 92.55 & 16.94 & 7.59 \\

        & \cellcolor{blue!8}{Doubao-Seed-2.0-Mini}
        & \cellcolor{blue!8}{\textcolor{red}{\textbf{91.31}}}
        & \cellcolor{blue!8}{\textcolor{red}{\textbf{86.97}}}
        & \cellcolor{blue!8}{\textcolor{red}{\textbf{94.46}}}
        & \cellcolor{blue!8}{\textcolor{red}{\textbf{21.26}}}
        & \cellcolor{blue!8}{\textcolor{red}{\textbf{23.45}}} \\

        & \cellcolor{blue!8}{DeepSeek-V4-Flash}
        & \cellcolor{blue!8}{90.26}
        & \cellcolor{blue!8}{\textcolor{blue}{\textbf{86.94}}}
        & \cellcolor{blue!8}{\textcolor{blue}{\textbf{93.22}}}
        & \cellcolor{blue!8}{\textcolor{blue}{\textbf{18.92}}}
        & \cellcolor{blue!8}{18.62} \\

        & \cellcolor{blue!8}{GPT-5.6-Terra}
        & \cellcolor{blue!8}{\textcolor{blue}{\textbf{90.53}}}
        & \cellcolor{blue!8}{86.82}
        & \cellcolor{blue!8}{93.04}
        & \cellcolor{blue!8}{17.32}
        & \cellcolor{blue!8}{\textcolor{blue}{\textbf{22.76}}} \\

        \midrule

        \multirow{3}{*}{DeepSeek-V4-Flash}
        & --
        & 84.92
        & 78.90
        & 96.64
        & 53.41
        & 55.86 \\

        & \cellcolor{blue!8}{Qwen3-4B-Instruct}
        & \cellcolor{blue!8}{\textcolor{blue}{\textbf{89.08}}}
        & \cellcolor{blue!8}{\textcolor{blue}{\textbf{81.59}}}
        & \cellcolor{blue!8}{\textcolor{blue}{\textbf{96.70}}}
        & \cellcolor{blue!8}{\textcolor{red}{\textbf{60.22}}}
        & \cellcolor{blue!8}{\textcolor{red}{\textbf{62.07}}} \\

        & \cellcolor{blue!8}{Qwen2.5-7B-Instruct}
        & \cellcolor{blue!8}{\textcolor{red}{\textbf{89.77}}}
        & \cellcolor{blue!8}{\textcolor{red}{\textbf{82.15}}}
        & \cellcolor{blue!8}{\textcolor{red}{\textbf{96.79}}}
        & \cellcolor{blue!8}{\textcolor{blue}{\textbf{60.13}}}
        & \cellcolor{blue!8}{\textcolor{blue}{\textbf{61.38}}} \\

        \bottomrule
    \end{tabular}
    }
\end{table}

\subsection{Ablation Studies}
We conduct ablation studies to evaluate the contributions of each component in SkillEarth.
As shown in Tab.~\ref{tab:ablation_main}, the baseline refers to directly extracted skills without subsequent evolution. 
Introducing the \emph{Tool Judge} improves tool execution reliability by validating tool selection and argument names.
Further incorporating the \emph{Semantic Judge} provides task-level semantic validation and feedback for skill refinement, leading to improved performance.
Building on these two components, the GOR graph explicitly captures inter-tool dependencies, enabling more precise and targeted skill refinement.
With all three components, SkillEarth achieves the best results across all five metrics.
Complete ablation results and detailed analyses are provided in \textbf{Supp.}~\ref{sec:supp_ablation}.

%% file: sec/5_conclusion.tex
\section{Conclusion}
In this paper, we proposed SkillEarth, a self-evolving skill extraction framework that equips general-purpose agents with reusable RS tool-call skills without expert intervention or parameter updates.
Unlike existing skill extraction methods that rely on holistic skill refinement, SkillEarth decomposes skills into independently evolvable components and introduces observation-grounded evolution.
By combining dual-judge feedback with grounded observation routing graphs, SkillEarth enables targeted refinement based on real tool execution evidence.
Experiments on two benchmarks demonstrate its effectiveness, outperforming existing RS agents and skill extraction methods across most metrics.
Furthermore, the extracted skills generalize across diverse LLMs and support cross-model transfer without parameter updates.

%% file: sec/6_statement.tex
\section*{LLM Usage Statement}
In this work, OpenAI Codex (GPT-6 Astra) assists with code development, debugging, and manuscript preparation, including English language polishing, grammatical corrections, and LaTeX formatting. 
The use of LLMs as components of the proposed framework is described in the methodology and experimental setup. 
The authors review and verify all AI-assisted content and take full responsibility for the research and its conclusions.

%% file: sec/X_suppl.tex
\clearpage
\setcounter{page}{1}

\maketitlesupplementary


\begin{algorithm}[!t]
\caption{Pseudocode of \name}
\label{alg:skillearth}
\begin{algorithmic}[1]
    \Require Dataset $\mathcal{D}$, budgets $n_t,n_s$, acceptance threshold $\gamma$.
    \Ensure SkillBank $\mathcal{B}$

    \State Initialize SkillBank $\mathcal{B} \gets \emptyset$

    \For{each sample $\xi_i\in\mathcal{D}$}
        \Statex \hspace{\algorithmicindent}\textit{\textcolor{blue}{\# Step 1. Component-wise Skill Extraction}}
        \State Initialize a skill template via Eq.~\ref{deterministic}.
        \State Extract initial skill $s_i^{(1)}$ via Eqs.~\ref{extraction}--\ref{Overwrite}.
        \Statex \hspace{\algorithmicindent}\textit{\textcolor{blue}{\# Step 2. Observation-Grounded Skill Evolution}}
        \State $\Phi_{\mathrm{sem}}\gets\emptyset$, $\mathrm{accepted}\gets\mathrm{false}$, $\mathrm{toolPassed}\gets\mathrm{false}$
        \For{$g=1$ \textbf{to} $n_s$} \Comment{Semantic-judge rounds}
            \State $\Phi_{\mathrm{tool}}\gets\emptyset$
            \For{$k=1$ \textbf{to} $n_t$} \Comment{Tool-judge rounds}
            \State Obtain $\hat{\tau}_i^{(g,k)}$ and $\hat{y}_i^{(g,k)}$ by iterating Eq.~\ref{execution}.
            \State Obtain $b_{i,g,k}^{\mathrm{tool}}$ and $\phi_{i,g,k}^{\mathrm{tool}}$ via Eq.~\ref{tool_judge}.
            \If{$b_{i,g,k}^{\mathrm{tool}}=1$}
                \State $\mathrm{toolPassed}
                \gets\mathrm{true}$
                \State \textbf{break}
            \Else
                \State $\Phi_{\mathrm{tool}}
                \gets \Phi_{\mathrm{tool}}
                \cup\{\phi_{i,g,k}^{\mathrm{tool}}\}$
            \EndIf
            \EndFor

            \If{\textbf{not} $\mathrm{toolPassed}$}
                \State \textbf{break}
                \Comment{Tool validation exhausted}
            \EndIf
    
            \State Obtain $s_{i,g}^{\mathrm{sem}}$ and $\phi_{i,g}^{\mathrm{sem}}$ via Eq.~\ref{semantic_judge}.
            \If{($s_{i,g}^{\mathrm{sem}}>\gamma$) \textbf{or} (image exists)}
                \State $\hat{\tau}_i^\star\gets\hat{\tau}_i^{(g,k)}$, $\mathrm{accepted} \gets \mathrm{true}$
                \State \textbf{break}
            \Else
            \State $\Phi_{\mathrm{sem}} \gets \Phi_{\mathrm{sem}} \cup \{\phi_{i,g}^{\mathrm{sem}}\}$
            \EndIf
        \EndFor

        \If{\textbf{not} $\mathrm{accepted}$}
            \State \textbf{continue}
            \Comment{Discard current sample}
        \EndIf

        \State Build the GOR graph from the accepted trajectory.
        \State Refine the skill to obtain $s_i^{(2)}$ via Eqs.~\ref{refine1} and~\ref{refine2}.
        \Statex \hspace{\algorithmicindent}\textit{\textcolor{blue}{\# Step 3. Skill Decision}}
        \State Compute $C_{ij}$ via~\ref{matching}.
        \If{$C_{ij}=0$ for all $s_j \in \mathcal{B}$}
            \State Add $s_i^{(2)}$ to $\mathcal{B}$.
        \Else
            \State Obtain decision $d_{ij}$ via Eq.~\ref{decision}.
            \If{$d_{ij}=\mathrm{Add}$}
                \State Add $s_i^{(2)}$ to $\mathcal{B}$.
            \ElsIf{$d_{ij}=\mathrm{Merge}$}
                \State Obtain $s_{ij}^{*}$ via Eq.~\ref{merging} and Delete $s_j^*$.
            \Else
                \State Discard $s_i^{(2)}$.
            \EndIf
        \EndIf
    \EndFor
    \State \Return $\mathcal{B}$
\end{algorithmic}
\end{algorithm}

\section{Qualitative Analysis of Skill Evolution}
\label{sec:skill_evolution_case}

In Sec.~\ref{sec:ogse}, Fig.~\ref{fig:skill1} illustrates a representative case of skill evolution for forest burn severity and vegetation regrowth analysis. 
While the initial and evolved skills preserve the same task objective and tool set, they differ in task planning and grounded observation routing.

\textbf{Task planning.} As shown in Fig.~\ref{fig:skill1}(a), the initial skill contains an over-specified plan that prescribes tool invocations and their execution order, closely following the original trajectory: it specifies two \texttt{AddIndexLayer} calls followed by \texttt{ComputeIndexChange}. 
In contrast, the evolved skill in Fig.~\ref{fig:skill1}(b) abstracts these tool-specific instructions into general analytical steps, including boundary retrieval, NBR computation, change analysis, and result summarization. 
This decouples task planning from specific tool invocations, improving skill generalizability.

\textbf{Grounded observation routing.} The initial skill specifies inter-tool dependencies ambiguously, leaving the mapping between tool outputs and downstream arguments unclear.
The evolved skill explicitly defines these mappings:
the \texttt{gpkg} output of \texttt{GetAreaBoundary} is forwarded to downstream tools.
The outputs of two \texttt{AddIndexLayer} calls are assigned to the \texttt{layer1name} and \texttt{layer2name} arguments of \texttt{ComputeIndexChange}, respectively.
The resulting \texttt{text\_output} is then passed to \texttt{Terminate} to generate the final response.
These explicit mappings reduce ambiguity and improve execution reliability.

\section{Algorithmic Pseudocode of \name}
\label{sec:algorithm}
We present the complete workflow of \name for SkillBank construction in Alg.~\ref{alg:skillearth}, including component-wise skill extraction, observation-grounded skill evolution, and skill decision.
It summarizes the key operations and control flow of the proposed framework.

\section{Implementation and Evaluation Details}
\label{supp_sec:implementation}

\begin{table*}[t]
\centering
\caption{Training-set composition after trajectory-stratified sampling. ``Traj.'' denotes the number of distinct tool-call trajectories within a task type. For each type--trajectory combination, up to 10 samples were retained. Original denotes the number of available training samples in each task type, while Selected denotes the number of samples used for SkillBank construction.}
\label{tab:oea-dataset}
\small
\setlength{\tabcolsep}{0.7pt}
\renewcommand{\arraystretch}{0.9}
\begin{tabular}{@{}lrrr@{\hspace{1.2em}}lrrr@{\hspace{1.2em}}lrrr@{}}
\toprule
\multicolumn{4}{c}{\textbf{Task types}} & \multicolumn{4}{c}{\textbf{Task types (continued)}} & \multicolumn{4}{c}{\textbf{Task types (continued)}} \\
\cmidrule(lr){1-4}\cmidrule(lr){5-8}\cmidrule(l){9-12}
\textbf{Task type} & \textbf{Traj.} & \textbf{Original} & \textbf{Selected} &
\textbf{Task type} & \textbf{Traj.} & \textbf{Original} & \textbf{Selected} &
\textbf{Task type} & \textbf{Traj.} & \textbf{Original} & \textbf{Selected} \\
\midrule
$\langle\mathrm{missing}\rangle$ & 5 & 13 & 13 & complex\_type3 & 1 & 21 & 10 & type14\_SAR & 1 & 87 & 10 \\
b1\_type1 & 1 & 65 & 10 & complex\_type4 & 1 & 46 & 10 & type15 & 1 & 99 & 10 \\
b1\_type2 & 1 & 63 & 10 & complex\_type5 & 1 & 9 & 9 & type19 & 3 & 395 & 30 \\
b1\_type3 & 1 & 63 & 10 & complex\_type6 & 1 & 9 & 9 & type1\_SAR & 3 & 110 & 30 \\
b1\_type4 & 1 & 58 & 10 & complex\_type7 & 1 & 1 & 1 & type25 & 4 & 739 & 40 \\
b1\_type5 & 1 & 62 & 10 & complex\_type8 & 1 & 13 & 10 & type25\_SAR & 4 & 101 & 40 \\
b1\_type6 & 1 & 59 & 10 & complex\_type9 & 1 & 9 & 9 & type30 & 1 & 2064 & 10 \\
b1\_type7 & 1 & 62 & 10 & gis\_type1 & 1 & 873 & 10 & type31 & 13 & 2176 & 79 \\
b1\_type8 & 1 & 62 & 10 & gis\_type2 & 2 & 394 & 10 & type32 & 3 & 251 & 30 \\
b2\_type1 & 1 & 29 & 10 & gis\_type3 & 1 & 87 & 10 & type38 & 1 & 100 & 10 \\
b2\_type2 & 1 & 29 & 10 & gis\_type4 & 1 & 683 & 10 & type39 & 1 & 100 & 10 \\
b2\_type3 & 1 & 29 & 10 & gis\_type5 & 2 & 393 & 11 & type4 & 5 & 764 & 50 \\
b2\_type4 & 1 & 28 & 10 & gis\_type6 & 2 & 317 & 11 & type40 & 5 & 92 & 45 \\
b2\_type5 & 1 & 28 & 10 & ind\_nbr & 2 & 75 & 20 & type41 & 1 & 149 & 10 \\
b2\_type6 & 1 & 20 & 10 & ind\_ndbi & 2 & 108 & 20 & type42 & 4 & 11 & 11 \\
b2\_type7 & 1 & 28 & 10 & ind\_ndvi & 2 & 40 & 13 & type4\_SAR & 3 & 112 & 30 \\
b2\_type8 & 1 & 28 & 10 & type1 & 3 & 884 & 30 & type5 & 3 & 763 & 30 \\
complex\_type1 & 1 & 33 & 10 & type11 & 6 & 122 & 35 & type5\_SAR & 3 & 112 & 30 \\
complex\_type10 & 1 & 23 & 10 & type13 & 3 & 74 & 28 & type8 & 4 & 72 & 37 \\
complex\_type2 & 1 & 9 & 9 & type14 & 1 & 1165 & 10 & type9 & 3 & 127 & 30 \\
\midrule
\rowcolor{blue!8}
\multicolumn{3}{r}{\textbf{Total}} &
\textbf{14538} &
\multicolumn{7}{>{\columncolor{blue!8}[1.2em][\tabcolsep]}r}{\textbf{Selected total}} &
\textbf{1050} \\
\bottomrule
\end{tabular}%
\end{table*}


\subsection{Dataset Construction}
\label{supp_sec:dataset}
As discussed in Sec.~\ref{sec:4.1}, we evaluate \name on OpenEarth-Bench~\cite{OpenEarthAgentSFT} and Earth-Bench~\cite{EarthAgentEO}.
OpenEarth-Bench provides 14,538 training trajectories and 24 tools spanning five categories: Perceptual Tools, GIS Computation Tools, Spectral Tools, Georeferenced Raster Tools, and Utility Tools.
These trajectories are originally designed for supervised fine-tuning of LLMs to acquire tool-calling capabilities.
In contrast, \name aims to extract reusable skills from limited execution experience without model fine-tuning.
However, since skill evolution involves actual tool execution, processing all 14,538 trajectories using Qwen3-4B-Instruct alone is estimated to take approximately 34.4 days.
To reduce this overhead while maintaining task diversity, we select the first 10 samples for each distinct tool-calling trajectory within each task type (or all available samples if fewer than 10), yielding 1,050 trajectories (7.2\% of the training set) for SkillBank construction.
This reduces the construction time to approximately 2 days.
For fair comparison, all existing skill extraction methods use the same subset for skill extraction.
Moreover, all compared methods, including RS agents and skill extraction methods, are evaluated on the full set of 1,169 evaluation instances.
The detailed distribution of the selected subset is provided in Tab.~\ref{tab:oea-dataset}.




\begin{table*}[t]
\centering
\caption{LLM configurations used in SkillEarth experiments. Temp. denotes the temperature parameter of LLMs.}
\label{tab:llm_config}
\small
\setlength{\tabcolsep}{3pt}
\renewcommand{\arraystretch}{0.9}
\begin{tabular}{clccc}
\toprule
\textbf{Category} & \textbf{Backbone} & \textbf{Usage in \name} & 
\textbf{Temp.} & \textbf{Max Tokens} \\
\midrule

\multirow{7}{*}{\makecell{Experimental\\LLMs}}
& GPT-5.6-Terra~\cite{openai2026gpt56} 
& \multirow{7}{*}{\makecell{\emph{Extractor}, \emph{Executor},\\\emph{Refiner}, \emph{Decider}, \emph{Merger}}}
& \multirow{7}{*}{0.0}
& \multirow{7}{*}{\makecell{1024 / 16384$^{\dagger}$}} \\

& Doubao-Seed-2.0-Mini~\cite{bytedance2026seed2} & & & \\
& DeepSeek-V4-Flash~\cite{deepseek2026v4} & & & \\
& Mistral-v0.3-Instruct~\cite{mistralai2024mistral7b} & & & \\
& DeepSeek-R1~\cite{deepseek2025r1} & & & \\
& Qwen2.5-7B-Instruct~\cite{qwen2024qwen25} & & & \\
& Qwen3-4B-Instruct~\cite{qwen2025qwen3} & & & \\

\midrule

\makecell{Cloud-based\\Judge}
& Doubao-Seed-Evolving~\cite{doubao_seed_evolving}
& \emph{Tool/Semantic Judge}
& 0.0
& 700 \\

\bottomrule
\end{tabular}

{\footnotesize
$^{\dagger}$1024 and 16384 denote the maximum output tokens for OpenEarth-Bench and Earth-Bench, respectively.
}
\end{table*}

\subsection{Model Configurations}
We summarize the model configurations used in \name in Tab.~\ref{tab:llm_config}.
We instantiate all five roles, including \emph{Extractor}, \emph{Executor}, \emph{Refiner}, \emph{Decider}, and \emph{Merger}, with multiple representative LLM backbones.
Specifically, we evaluate seven LLMs, including three frontier LLMs (GPT-5.6-Terra~\cite{openai2026gpt56}, Doubao-Seed-2.0-Mini~\cite{bytedance2026seed2}, and DeepSeek-V4-Flash~\cite{deepseek2026v4}) and four open-source LLMs (Mistral-v0.3-Instruct~\cite{mistralai2024mistral7b}, DeepSeek-R1~\cite{deepseek2025r1}, Qwen2.5-7B-Instruct~\cite{qwen2024qwen25}, and Qwen3-4B-Instruct~\cite{qwen2025qwen3}).
We employ Doubao-Seed-Evolving~\cite{doubao_seed_evolving} as the cloud-based judge for both tool execution and semantic assessment.
All LLMs are configured with a temperature of 0.0 to ensure deterministic outputs and improve the reproducibility of skill extraction and tool execution. For different benchmarks, we set different maximum output token limits.
For OpenEarth-Bench, we use a maximum of 1,024 output tokens.
For Earth-Bench, we increase this limit to 16,384 tokens because its tasks often involve processing multiple files, maintaining long execution contexts, and generating intermediate results with absolute
file paths. 
The larger token budget prevents premature truncation of tool arguments, execution descriptions, and final responses, thereby ensuring complete task execution.

The hyperparameters for SkillBank construction and skill retrieval are listed in Table~\ref{tab:hyperparameters}.
We allow at most five tool-execution rounds ($n_t$) and two semantic refinement rounds ($n_s$).
A generated skill is accepted only when it satisfies the corresponding quality criterion: the semantic score exceeds 0.6 for text-based outputs, or the required image output is successfully generated ($\mathrm{exist}=1$) for image-based tasks. 
For skill retrieval, we encode skills using a hashing-based encoder with a 2048-dimensional representation and measure relevance using cosine similarity. 
For each query, only the top-1 skill ($k=1$) is retrieved and provided to the agent, avoiding excessive contextual interference while preserving the most relevant prior experience.

\subsection{Evaluation Metrics}
\subsubsection{Metrics for OpenEarth-Bench}
OpenEarth-Bench~\cite{OpenEarthAgentSFT} provides two evaluation protocols: Step-by-Step and End-to-End, assessing tool-calling behavior and complete task execution, respectively.

\begin{table}[t]
\centering
\caption{Hyperparameter settings for SkillBank construction and retrieval.
Here, $\mathrm{exist}=1$ indicates that the required image output is successfully generated.}
\label{tab:hyperparameters}
\small
\setlength{\tabcolsep}{2pt}
\renewcommand{\arraystretch}{0.9}
\begin{tabular}{clr}
\toprule
\textbf{Stage} & \textbf{Parameter} & \textbf{Setting} \\
\midrule

\multirow{4}{*}{\makecell{SkillBank\\Construction}}
& Max. tool rounds ($n_t$) & 5 \\
& Max. semantic rounds ($n_s$) & 2 \\
& \multirow{2}{*}{Acceptance rule} 
& Text: $s>0.6$ \\
& & Image: $\mathrm{exist}=1$ \\

\midrule

\multirow{4}{*}{\makecell{Skill\\Retrieval}}
& Encoder & Hashing \\
& Dimension & 2048 \\
& Similarity & Cosine \\
& Top-$k$ & 1 \\

\bottomrule
\end{tabular}
\end{table}

\noindent\textbf{Step-by-Step Evaluation.}
This protocol measures procedural reasoning and tool invocation behavior.
\begin{itemize}
    \item \textbf{Instance Accuracy (Inst.):} Measures the proportion of tool calls without logical or syntactic errors.
    \item \textbf{Tool Accuracy (Tool.):} Evaluates the correctness of tool selection.
    \item \textbf{Argument Name Accuracy (ArgN.):} Checks whether all required arguments are provided.
    \item \textbf{Argument Value Accuracy (ArgV.):} Assesses the correctness of argument values.
    \item \textbf{Summarization Accuracy (Summ.):} Evaluates the quality of final responses synthesized from previous tools.
\end{itemize}

\noindent\textbf{End-to-End Evaluation.}
This protocol evaluates autonomous task execution with live tool use.
\begin{itemize}
    \item \textbf{Tool F1:} Measures tool selection using F1 scores across four functional categories: Perception (Per.), Operation (Op.), Logic (Logic.), and GIS (GIS.).
    
    \item \textbf{Tool Order:} Evaluates trajectory correctness at three levels: Unique (Uni.) checks tool correspondence while ignoring repeated calls; AnyOrder (AnyOr.) requires all tool calls, including duplicates, regardless of order, and SameOrder (SameO.) additionally enforces the correct invocation sequence.
    
    \item \textbf{Task Accuracy:} Evaluates final outputs against reference answers. Generation Accuracy (Gen.) assesses image generation tasks based on valid generation tool calls and successful execution, while Answer Accuracy (Ans.) evaluates non-generation tasks using seedance-2.0-mini with a standardized evaluation prompt.

\end{itemize}

\begin{table*}[t]
    \centering
    \small
    \setlength{\tabcolsep}{2.0pt}
    \setlength{\fboxsep}{0.5pt}
    \renewcommand{\arraystretch}{0.78}
    \caption{Step-by-Step cross-model skill transfer on OpenEarth-Bench. ``Skill Source'' denotes the model used to extract the transferred skills. The best, second-best, and third-best results within each evaluation model are highlighted in \textcolor{red}{\textbf{red}}, \textcolor{blue}{\textbf{blue}}, and \textcolor{orange}{\textbf{orange}}, respectively.}
    \label{tab:transfer_step}
    \begin{tabular}{lcl*{5}{c}}
        \toprule
        \textbf{Evaluation Model}
        & \textbf{Setting}
        & \textbf{Skill Source}
        & \textbf{Inst.} $\uparrow$
        & \textbf{Tool.} $\uparrow$
        & \textbf{ArgN.} $\uparrow$
        & \textbf{ArgV.} $\uparrow$
        & \textbf{Summ.} $\uparrow$ \\
        \midrule
        \multirow{4}{*}{Qwen3-4B-Instruct}
        & w/o skill
        & \multicolumn{1}{c}{--}
        & \textcolor{orange}{\textbf{98.47}}
        & 87.70 & 84.33 & 33.67 & 92.55 \\
        & \cellcolor{blue!8}{w/ skill}
        & \cellcolor{blue!8}{Doubao-Seed-2.0-Mini}
        & \cellcolor{blue!8}{\textcolor{blue}{\textbf{98.66}}}
        & \cellcolor{blue!8}{\textcolor{red}{\textbf{91.31}}}
        & \cellcolor{blue!8}{\textcolor{red}{\textbf{86.97}}}
        & \cellcolor{blue!8}{\textcolor{red}{\textbf{37.12}}}
        & \cellcolor{blue!8}{\textcolor{red}{\textbf{94.46}}} \\
        & \cellcolor{blue!8}{w/ skill}
        & \cellcolor{blue!8}{DeepSeek-V4-Flash}
        & \cellcolor{blue!8}{\textcolor{orange}{\textbf{98.47}}}
        & \cellcolor{blue!8}{\textcolor{orange}{\textbf{90.26}}}
        & \cellcolor{blue!8}{\textcolor{blue}{\textbf{86.94}}}
        & \cellcolor{blue!8}{\textcolor{blue}{\textbf{37.07}}}
        & \cellcolor{blue!8}{\textcolor{blue}{\textbf{93.22}}} \\
        & \cellcolor{blue!8}{w/ skill}
        & \cellcolor{blue!8}{GPT-5.6-Terra}
        & \cellcolor{blue!8}{\textcolor{red}{\textbf{98.80}}}
        & \cellcolor{blue!8}{\textcolor{blue}{\textbf{90.53}}}
        & \cellcolor{blue!8}{\textcolor{orange}{\textbf{86.82}}}
        & \cellcolor{blue!8}{\textcolor{orange}{\textbf{36.20}}}
        & \cellcolor{blue!8}{\textcolor{orange}{\textbf{93.04}}} \\
        \cmidrule(lr){2-8}
        \multirow{3}{*}{DeepSeek-V4-Flash}
        & w/o skill
        & \multicolumn{1}{c}{--}
        & \textcolor{orange}{\textbf{99.78}}
        & \textcolor{orange}{\textbf{84.92}}
        & \textcolor{orange}{\textbf{78.90}}
        & \textcolor{orange}{\textbf{40.32}}
        & \textcolor{orange}{\textbf{96.64}} \\
        & \cellcolor{blue!8}{w/ skill}
        & \cellcolor{blue!8}{Qwen3-4B-Instruct}
        & \cellcolor{blue!8}{\textcolor{blue}{\textbf{99.92}}}
        & \cellcolor{blue!8}{\textcolor{blue}{\textbf{89.08}}}
        & \cellcolor{blue!8}{\textcolor{blue}{\textbf{81.59}}}
        & \cellcolor{blue!8}{\textcolor{red}{\textbf{43.41}}}
        & \cellcolor{blue!8}{\textcolor{blue}{\textbf{96.70}}} \\
        & \cellcolor{blue!8}{w/ skill}
        & \cellcolor{blue!8}{Qwen2.5-7B-Instruct}
        & \cellcolor{blue!8}{\textcolor{red}{\textbf{99.98}}}
        & \cellcolor{blue!8}{\textcolor{red}{\textbf{89.77}}}
        & \cellcolor{blue!8}{\textcolor{red}{\textbf{82.15}}}
        & \cellcolor{blue!8}{\textcolor{blue}{\textbf{43.24}}}
        & \cellcolor{blue!8}{\textcolor{red}{\textbf{96.79}}} \\
        \bottomrule
    \end{tabular}
\end{table*}

\begin{table*}[t]
    \centering
    \small
    \setlength{\tabcolsep}{1.2pt}
    \setlength{\fboxsep}{0.5pt}
    \renewcommand{\arraystretch}{0.78}
    \caption{End-to-End cross-model skill transfer on OpenEarth-Bench. ``Skill Source'' denotes the model used to extract the transferred skills. The best, second-best, and third-best results within each evaluation model are highlighted in \textcolor{red}{\textbf{red}}, \textcolor{blue}{\textbf{blue}}, and \textcolor{orange}{\textbf{orange}}, respectively.}
    \label{tab:transfer_e2e}
    \begin{tabular}{lcl*{9}{c}}
        \toprule
        \multirow{2}{*}{\textbf{Evaluation Model}}
        & \multirow{2}{*}{\textbf{Setting}}
        & \multirow{2}{*}{\textbf{Skill Source}}
        & \multicolumn{4}{c}{\textbf{F$_1$ Scores}}
        & \multicolumn{3}{c}{\textbf{Tool Order}}
        & \multicolumn{2}{c}{\textbf{Accuracy}} \\
        \cmidrule(lr){4-7}
        \cmidrule(lr){8-10}
        \cmidrule(lr){11-12}
        & & &
        \textbf{Per.} $\uparrow$
        & \textbf{Op.} $\uparrow$
        & \textbf{Logic.} $\uparrow$
        & \textbf{GIS.} $\uparrow$
        & \textbf{AnyOr.} $\uparrow$
        & \textbf{SameO.} $\uparrow$
        & \textbf{Uni.} $\uparrow$
        & \textbf{Ans.} $\uparrow$
        & \textbf{Gen.} $\uparrow$ \\
        \midrule
        \multirow{4}{*}{Qwen3-4B-Instruct}
        & w/o skill
        & \multicolumn{1}{c}{--}
        & 13.31 & 12.54 & 11.80
        & \textcolor{orange}{\textbf{76.42}}
        & 15.23 & 14.63 & 18.48
        & 16.94 & 7.59 \\
        & \cellcolor{blue!8}{w/ skill}
        & \cellcolor{blue!8}{Doubao-Seed-2.0-Mini}
        & \cellcolor{blue!8}{\textcolor{red}{\textbf{33.14}}}
        & \cellcolor{blue!8}{\textcolor{red}{\textbf{33.58}}}
        & \cellcolor{blue!8}{\textcolor{red}{\textbf{23.82}}}
        & \cellcolor{blue!8}{\textcolor{red}{\textbf{80.23}}}
        & \cellcolor{blue!8}{\textcolor{red}{\textbf{23.95}}}
        & \cellcolor{blue!8}{\textcolor{red}{\textbf{23.78}}}
        & \cellcolor{blue!8}{\textcolor{red}{\textbf{27.20}}}
        & \cellcolor{blue!8}{\textcolor{red}{\textbf{21.26}}}
        & \cellcolor{blue!8}{\textcolor{red}{\textbf{23.45}}} \\
        & \cellcolor{blue!8}{w/ skill}
        & \cellcolor{blue!8}{DeepSeek-V4-Flash}
        & \cellcolor{blue!8}{\textcolor{orange}{\textbf{23.98}}}
        & \cellcolor{blue!8}{\textcolor{orange}{\textbf{25.58}}}
        & \cellcolor{blue!8}{\textcolor{blue}{\textbf{20.09}}}
        & \cellcolor{blue!8}{\textcolor{blue}{\textbf{77.53}}}
        & \cellcolor{blue!8}{\textcolor{blue}{\textbf{22.07}}}
        & \cellcolor{blue!8}{\textcolor{blue}{\textbf{21.73}}}
        & \cellcolor{blue!8}{\textcolor{blue}{\textbf{24.12}}}
        & \cellcolor{blue!8}{\textcolor{blue}{\textbf{18.92}}}
        & \cellcolor{blue!8}{\textcolor{orange}{\textbf{18.62}}} \\
        & \cellcolor{blue!8}{w/ skill}
        & \cellcolor{blue!8}{GPT-5.6-Terra}
        & \cellcolor{blue!8}{\textcolor{blue}{\textbf{29.26}}}
        & \cellcolor{blue!8}{\textcolor{blue}{\textbf{31.69}}}
        & \cellcolor{blue!8}{\textcolor{orange}{\textbf{18.48}}}
        & \cellcolor{blue!8}{76.21}
        & \cellcolor{blue!8}{\textcolor{orange}{\textbf{20.62}}}
        & \cellcolor{blue!8}{\textcolor{orange}{\textbf{20.36}}}
        & \cellcolor{blue!8}{\textcolor{orange}{\textbf{22.93}}}
        & \cellcolor{blue!8}{\textcolor{orange}{\textbf{17.32}}}
        & \cellcolor{blue!8}{\textcolor{blue}{\textbf{22.76}}} \\
        \cmidrule(lr){2-12}
        \multirow{3}{*}{DeepSeek-V4-Flash}
        & w/o skill
        & \multicolumn{1}{c}{--}
        & \textcolor{orange}{\textbf{26.05}}
        & \textcolor{red}{\textbf{38.51}}
        & \textcolor{orange}{\textbf{13.11}}
        & \textcolor{blue}{\textbf{82.27}}
        & \textcolor{orange}{\textbf{50.90}}
        & \textcolor{orange}{\textbf{50.21}}
        & \textcolor{orange}{\textbf{56.54}}
        & \textcolor{orange}{\textbf{53.41}}
        & \textcolor{orange}{\textbf{55.86}} \\
        & \cellcolor{blue!8}{w/ skill}
        & \cellcolor{blue!8}{Qwen3-4B-Instruct}
        & \cellcolor{blue!8}{\textcolor{red}{\textbf{40.84}}}
        & \cellcolor{blue!8}{\textcolor{blue}{\textbf{37.60}}}
        & \cellcolor{blue!8}{\textcolor{blue}{\textbf{31.91}}}
        & \cellcolor{blue!8}{\textcolor{orange}{\textbf{81.79}}}
        & \cellcolor{blue!8}{\textcolor{red}{\textbf{57.74}}}
        & \cellcolor{blue!8}{\textcolor{red}{\textbf{57.31}}}
        & \cellcolor{blue!8}{\textcolor{red}{\textbf{62.70}}}
        & \cellcolor{blue!8}{\textcolor{red}{\textbf{60.22}}}
        & \cellcolor{blue!8}{\textcolor{red}{\textbf{62.07}}} \\
        & \cellcolor{blue!8}{w/ skill}
        & \cellcolor{blue!8}{Qwen2.5-7B-Instruct}
        & \cellcolor{blue!8}{\textcolor{blue}{\textbf{40.67}}}
        & \cellcolor{blue!8}{\textcolor{orange}{\textbf{35.96}}}
        & \cellcolor{blue!8}{\textcolor{red}{\textbf{36.18}}}
        & \cellcolor{blue!8}{\textcolor{red}{\textbf{82.65}}}
        & \cellcolor{blue!8}{\textcolor{blue}{\textbf{56.72}}}
        & \cellcolor{blue!8}{\textcolor{blue}{\textbf{56.29}}}
        & \cellcolor{blue!8}{\textcolor{blue}{\textbf{61.93}}}
        & \cellcolor{blue!8}{\textcolor{blue}{\textbf{60.13}}}
        & \cellcolor{blue!8}{\textcolor{blue}{\textbf{61.38}}}
        \\
        \bottomrule
    \end{tabular}
\end{table*}

\begin{table}[t]
    \centering
    \caption{Ablation study of SkillEarth on OpenEarth-Bench under the Step-by-Step setting using Qwen3-4B-Instruct. 
    The top three results are highlighted in \textcolor{red}{red}, \textcolor{blue}{blue}, and \textcolor{orange}{orange}, respectively.}
    \label{tab:ablation_step}
    {
    \small
    \setlength{\tabcolsep}{0.6pt}
    \setlength{\fboxsep}{0.6pt}
    \renewcommand{\arraystretch}{0.9}
    \begin{tabular}{l c c c c c}
        \toprule
        \textbf{Setting} & \textbf{Ins. $\uparrow$} & \textbf{Tool. $\uparrow$} & \textbf{ArgN. $\uparrow$} & \textbf{ArgV. $\uparrow$} & \textbf{Summ. $\uparrow$} \\
        \midrule

        Baseline
        & \textcolor{blue}{\textbf{98.91}}
        & 90.33
        & 85.36
        & \textcolor{orange}{\textbf{35.66}}
        & 92.01 \\

        \quad + Tool Judge
        & \textcolor{orange}{\textbf{98.85}}
        & \textcolor{blue}{\textbf{90.92}}
        & \textcolor{orange}{\textbf{85.70}}
        & 35.29
        & \textcolor{orange}{\textbf{92.51}} \\

        \quad + Semantic Judge
        & 98.76
        & \textcolor{orange}{\textbf{90.81}}
        & \textcolor{blue}{\textbf{85.99}}
        & \textcolor{blue}{\textbf{35.95}}
        & \textcolor{blue}{\textbf{92.66}} \\

        \rowcolor{blue!8}
        \quad + GOR Graph (\textit{Ours})
        & \textcolor{red}{\textbf{99.07}}
        & \textcolor{red}{\textbf{91.04}}
        & \textcolor{red}{\textbf{86.48}}
        & \textcolor{red}{\textbf{36.17}}
        & \textcolor{red}{\textbf{92.67}} \\

        \bottomrule
    \end{tabular}
    }
\end{table}

\begin{table*}[t]
    \centering
    \caption{Ablation study of SkillEarth on OpenEarth-Bench under the End-to-End setting using Qwen3-4B-Instruct. The top three results are highlighted in \textcolor{red}{red}, \textcolor{blue}{blue}, and \textcolor{orange}{orange}, respectively.}
    \label{tab:ablation_e2e}
    {
    \small
    \setlength{\tabcolsep}{2pt}
    \setlength{\fboxsep}{0.6pt}
    \renewcommand{\arraystretch}{0.9}
    \begin{tabular}{l c c c c c c c c c}
        \toprule
        \multirow{2}{*}{\textbf{Setting}}
        & \multicolumn{4}{c}{\textbf{F$_1$ scores}}
        & \multicolumn{3}{c}{\textbf{Tool Order}}
        & \multicolumn{2}{c}{\textbf{Accuracy}} \\
        \cmidrule(lr){2-5}
        \cmidrule(lr){6-8}
        \cmidrule(lr){9-10}
        & \textbf{Per. $\uparrow$}
        & \textbf{Op. $\uparrow$}
        & \textbf{Logic. $\uparrow$}
        & \textbf{GIS. $\uparrow$}
        & \textbf{AnyOr. $\uparrow$}
        & \textbf{SameO. $\uparrow$}
        & \textbf{Uni. $\uparrow$}
        & \textbf{Ans. $\uparrow$}
        & \textbf{Gen. $\uparrow$} \\
        \midrule

        Baseline
        & 31.71
        & 27.45
        & 24.38
        & 80.04
        & 23.61
        & 23.44
        & \textcolor{orange}{\textbf{26.60}}
        & 18.35
        & 18.62 \\

        \quad + Tool Judge
        & \textcolor{orange}{\textbf{32.43}}
        & \textcolor{red}{\textbf{36.44}}
        & \textcolor{orange}{\textbf{25.78}}
        & \textcolor{orange}{\textbf{81.02}}
        & \textcolor{orange}{\textbf{24.12}}
        & \textcolor{orange}{\textbf{23.78}}
        & \textcolor{orange}{\textbf{26.60}}
        & \textcolor{orange}{\textbf{18.42}}
        & \textcolor{orange}{\textbf{20.09}} \\

        \quad + Semantic Judge
        & \textcolor{blue}{\textbf{33.52}}
        & \textcolor{orange}{\textbf{31.58}}
        & \textcolor{blue}{\textbf{28.56}}
        & \textcolor{blue}{\textbf{81.24}}
        & \textcolor{blue}{\textbf{24.24}}
        & \textcolor{blue}{\textbf{24.15}}
        & \textcolor{blue}{\textbf{26.80}}
        & \textcolor{blue}{\textbf{18.49}}
        & \textcolor{blue}{\textbf{23.45}} \\

        \rowcolor{blue!8}
        \quad + GOR Graph (\textit{Ours})
        & \textcolor{red}{\textbf{35.00}}
        & \textcolor{blue}{\textbf{33.06}}
        & \textcolor{red}{\textbf{28.73}}
        & \textcolor{red}{\textbf{81.85}}
        & \textcolor{red}{\textbf{25.32}}
        & \textcolor{red}{\textbf{25.15}}
        & \textcolor{red}{\textbf{27.89}}
        & \textcolor{red}{\textbf{19.59}}
        & \textcolor{red}{\textbf{24.14}} \\

        \bottomrule
    \end{tabular}
    }
\end{table*}

\subsubsection{Metrics for Earth-Bench}
Earth-Bench~\cite{EarthAgentEO} supports two evaluation modes: Automatic Planning (AP), which assesses autonomous task planning, and Instruction Following (IF), which evaluates the execution of human-provided instructions.
Both modes employ Step-by-Step and End-to-End evaluations to assess intermediate reasoning and final answer, respectively.

\noindent\textbf{Step-by-Step Evaluation.}
This protocol evaluates intermediate reasoning trajectories through four metrics:
\begin{itemize}
    \item \textbf{Tool-Any-Order (AnyOr.):} Checks whether all required tools are selected, regardless of their invocation order.
    \item \textbf{Tool-In-Order (InOr.):} Evaluates whether tools are invoked in the correct sequence.
    \item \textbf{Tool-Exact-Match (Exact.):} Measures the exact prefix-level agreement between predicted and expert trajectories.
    \item \textbf{Parameter Accuracy (Param.):} Assesses whether tool identifiers and their arguments match the reference.
\end{itemize}

\noindent\textbf{End-to-End Evaluation.}
This protocol evaluates overall task performance using two metrics:
\begin{itemize}
    \item \textbf{Accuracy (Acc.):} Measures the correctness of answers.
    \item \textbf{Efficiency (Eff.):} Evaluates the efficiency of execution trajectories relative to expert solutions.
\end{itemize}


\begin{figure*}[t]
    \centering
    \includegraphics[width=\textwidth]{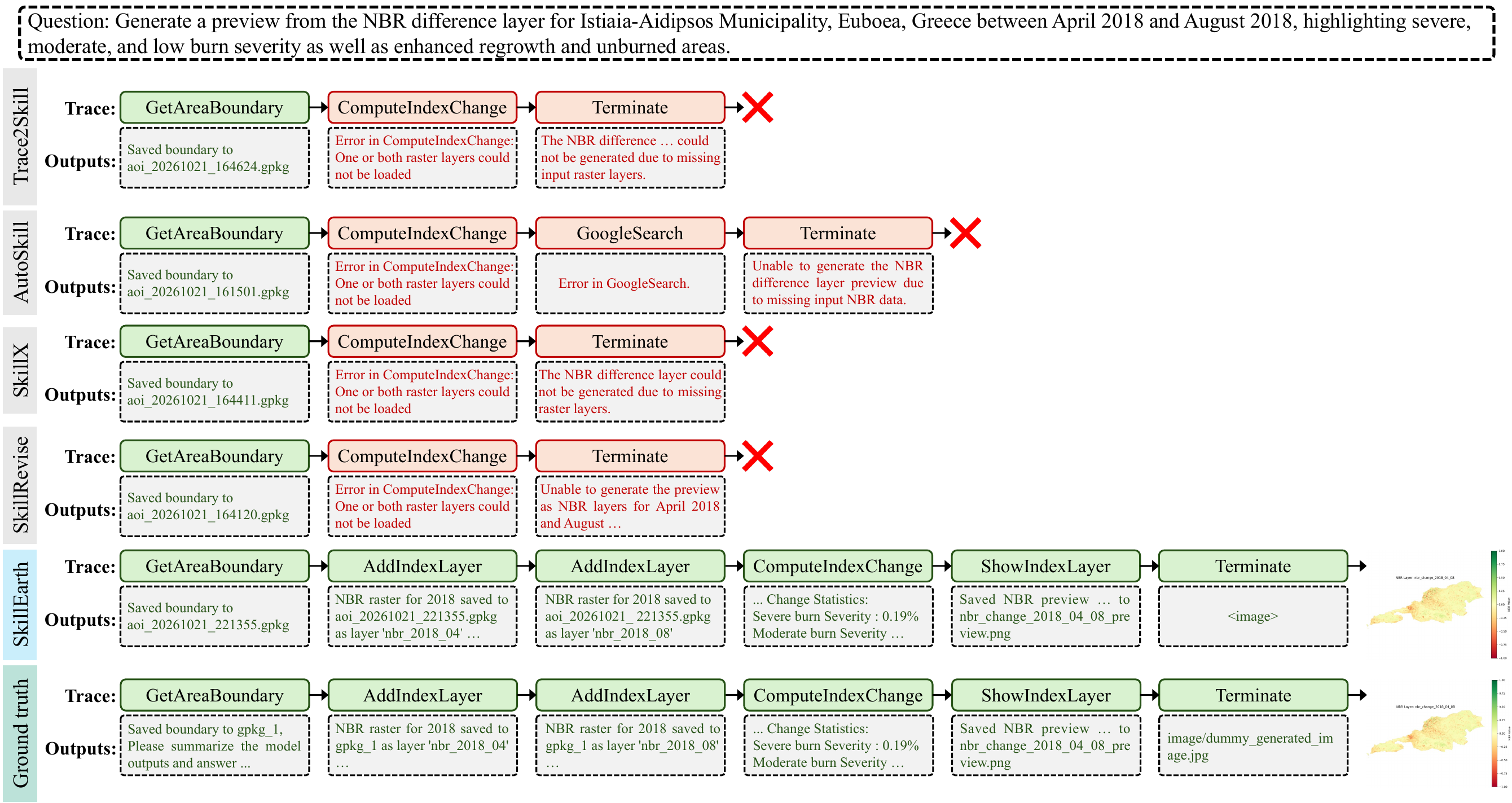}
    \caption{
    Qualitative comparison of \name with existing skill extraction methods. We visualize the tool-calling trajectories and outputs of Qwen3-4B-Instruct on an NBR-based burn severity analysis task. Existing methods fail due to missing raster layers, while \name successfully completes the task by following the correct tool execution sequence.
    }
    \label{fig:comparison_skill}
\end{figure*}

\section{Qualitative Comparison with RS Agents}
\label{sec:qual_rs}
Fig.~\ref{fig:comparison_rs} presents a qualitative comparison between \name and state-of-the-art RS agents. 
Specifically, we instruct different agents to identify the smallest airplane in an image and draw a bounding box around it, visualizing their tool-calling trajectories and corresponding tool outputs.
OpenEarthAgent invokes the \texttt{TextToBox} tool with the \texttt{Top1} parameter set to \texttt{True}, which returns only the bounding box with the highest confidence score. It then directly calls \texttt{DrawBox}, producing an incorrect result.
Earth-Agent, RSMeM, and RS-Agent first invoke \texttt{ObjectDetection} and successfully obtain all six bounding boxes. 
However, instead of using the \texttt{Calculator} tool to compute their areas and identify the smallest airplane, they directly select a bounding box without performing the necessary calculations. Consequently, Earth-Agent happens to select the correct box, whereas RSMeM and RS-Agent produce incorrect results.
In contrast, \name correctly invokes \texttt{TextToBox} with \texttt{Top1} set to \texttt{False}, retrieving all six bounding boxes. It then performs six \texttt{Calculator} calls to compute their respective areas, identifies the bounding box with the smallest area, and generates the correct image. 

The last row shows the ground-truth trajectory, which computes the areas of all six bounding boxes in a single \texttt{Calculator} call. 
In comparison, SkillEarth decomposes this operation into six separate calls, resulting in a more explicit and interpretable reasoning process.

\section{Qualitative Comparison with Skill Extraction Methods}
\label{sec:qual_skill}
Fig.~\ref{fig:comparison_skill} presents a qualitative comparison between \name and existing skill extraction methods. 
Specifically, we ask Qwen3-4B-Instruct to solve a burn-severity analysis task using the skills extracted by different methods. 
The task is to generate a preview from the NBR difference layer for Istiaia-Aidipsos Municipality, Euboea, Greece, between April 2018 and August 2018, highlighting severe, moderate, and low burn severity, as well as enhanced regrowth and unburned areas.

For the compared methods, \texttt{ComputeIndexChange} is invoked prematurely in the second step. 
However, the raster layers required by \texttt{ComputeIndexChange} are not obtained beforehand through \texttt{AddIndexLayer}, which causes tool failure and leads to task failure. 
In contrast, \name successfully obtains the target boundary, adds the raster layers for the start and end times, and then correctly invokes \texttt{ComputeIndexChange} and \texttt{ShowIndexLayer} to generate the final result image.

\begin{figure*}[t]
    \centering
    \includegraphics[width=\textwidth]{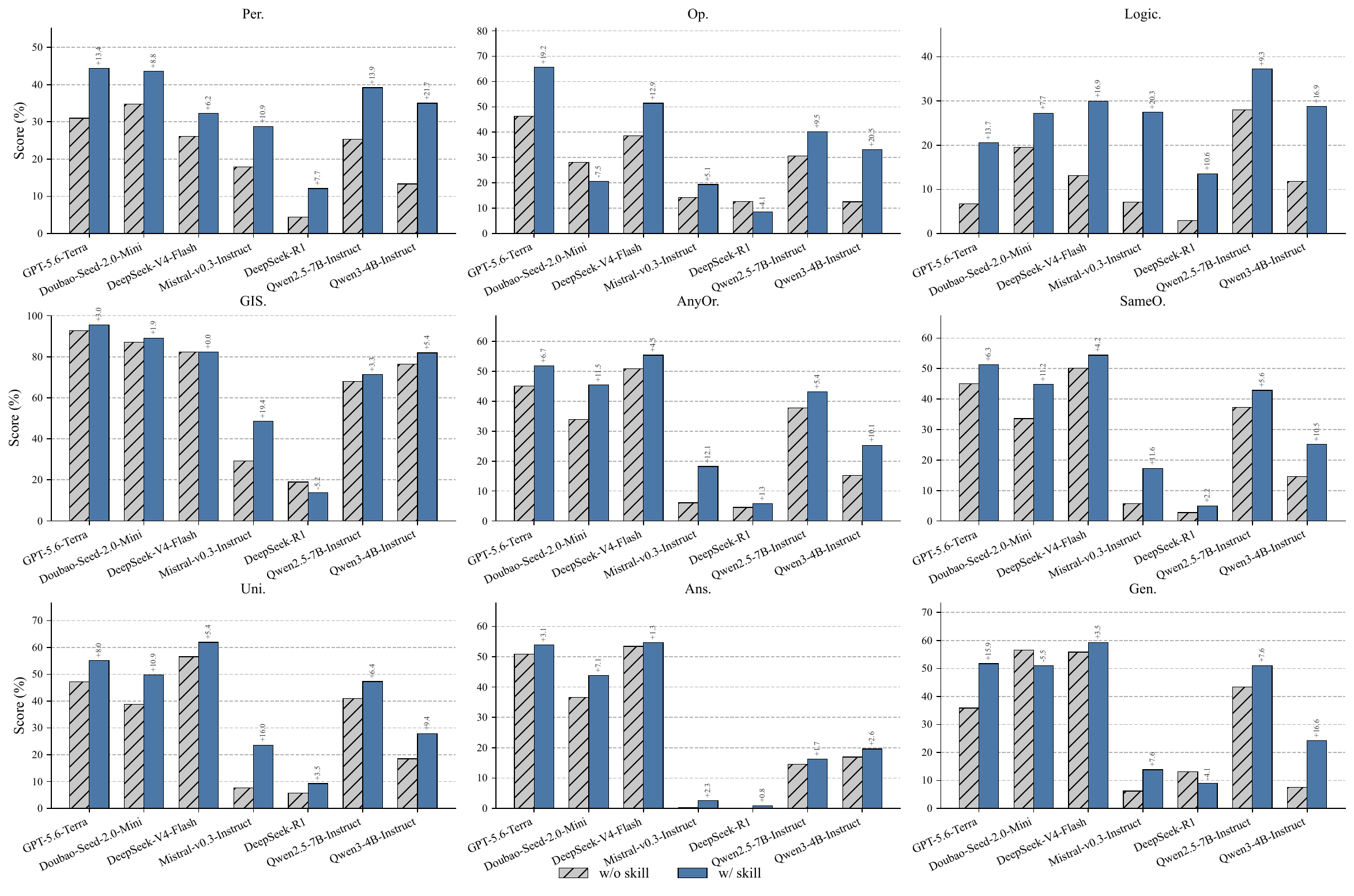}
    \caption{End-to-End generalization of \name across seven LLMs on OpenEarth-Bench. We compare performance with and without skill guidance across nine evaluation metrics. The values above the bars indicate performance changes in percentage points.}
    \label{fig:bar}
\end{figure*}

\section{Cross-Model Skill Transferability}
\label{transfer}
We evaluate the cross-model transferability of SkillEarth on OpenEarth-Bench.
As shown in Tabs.~\ref{tab:transfer_step} and~\ref{tab:transfer_e2e}, cross-model skill transfer generally improves both Step-by-Step and End-to-End performance, demonstrating the reusability of RS knowledge across different LLMs.

Transferring skills from frontier LLMs to Qwen3-4B-Instruct improves all five Step-by-Step metrics and most End-to-End metrics.
Conversely, transferring skills from Qwen3-4B-Instruct and Qwen2.5-7B-Instruct to DeepSeek-V4-Flash improves all five Step-by-Step metrics and seven and eight of the nine End-to-End metrics, respectively.
Notably, Qwen3-4B-Instruct skills increase Ans. from 53.41 to 60.22 and Gen. from 55.86 to 62.07, while Qwen2.5-7B-Instruct skills achieve a larger improvement in Logic. F$_1$ from 13.11 to 36.18.
These results demonstrate that skills extracted by smaller open-source LLMs can also enhance more capable frontier models.

\section{Detailed Ablation Analysis}
\label{sec:supp_ablation}
We analyze the contributions of each component in \name by progressively adding the Tool Judge, Semantic Judge, and GOR graph to the Qwen3-4B-Instruct baseline on OpenEarth-Bench.
Results under both Step-by-Step and End-to-End settings are reported in Tabs.~\ref{tab:ablation_e2e} and~\ref{tab:ablation_step}.

\textbf{Step-by-Step Evaluation.}
As shown in Tab.~\ref{tab:ablation_step}, the Tool Judge improves Tool. and ArgN. accuracy by reducing execution errors.
The Semantic Judge further improves ArgN. and Summ. accuracy despite a slight drop in Tool. accuracy, demonstrating the complementary role of semantic feedback.
Incorporating the GOR graph achieves the best results across all metrics, validating the effectiveness of \name.

\textbf{End-to-End Evaluation.}
As shown in Tab.~\ref{tab:ablation_e2e}, the Tool Judge improves task execution, while the Semantic Judge further enhances task-level performance.
Incorporating the GOR graph achieves the best results on eight of nine End-to-End metrics, including all three tool-order metrics (AnyOr., SameO., and Uni.).
These gains highlight the importance of explicit observation routing in preserving execution order and information flow, demonstrating the effectiveness of dual-judge feedback and GOR-based skill refinement for reliable multi-step RS workflows.

Overall, the three components play complementary roles: the Tool Judge validates execution, the Semantic Judge assesses task outcomes, and the GOR graph captures inter-tool dependencies.
Together, they enable SkillEarth to evolve reliable skills through execution-grounded evidence rather than outcome-based reflection alone.

\clearpage
\onecolumn
\section{System Prompts}
\label{supp_sec:prompts}

We provide the complete system prompts used in \name for reproducibility.

\subsection{Extractor}
\tcbinputlisting{
    promptstyle,
    title={System Prompt for the Extractor},
    listing file={prompt/extraction.txt}
}

\subsection{Executor}
\tcbinputlisting{
    promptstyle,
    title={System Prompt for the Executor},
    listing file={prompt/execution.txt}
}

\subsection{Refiner}
\tcbinputlisting{
    promptstyle,
    title={System Prompt for the Refiner (Grounded Observation Routing)},
    listing file={prompt/refinement_gor.txt}
}

\tcbinputlisting{
    promptstyle,
    title={System Prompt for the Refiner (Other Components)},
    listing file={prompt/refinement_others.txt}
}

\subsection{Decider}
\tcbinputlisting{
    promptstyle,
    title={System Prompt for the Decider},
    listing file={prompt/decision.txt}
}

\subsection{Merger}
\tcbinputlisting{
    promptstyle,
    title={System Prompt for the Merger},
    listing file={prompt/merge.txt}
}

\subsection{Tool Judge}
\tcbinputlisting{
    promptstyle,
    title={System Prompt for the Tool Judge},
    listing file={prompt/tool_judge.txt}
}

\subsection{Semantic Judge}
\tcbinputlisting{
    promptstyle,
    title={System Prompt for the Semantic Judge},
    listing file={prompt/semantic_judge.txt}
}

\clearpage
\twocolumn
